\documentclass[journal]{IEEEtran}
\usepackage{lineno}
\usepackage{amsmath,amssymb,amsfonts}
\usepackage{algorithmic}
\usepackage{graphicx}
\usepackage{textcomp}
\usepackage{xcolor}
\usepackage{booktabs}
\usepackage{subcaption}
\usepackage[hidelinks]{hyperref}
\usepackage{balance}
\usepackage{soul}
\usepackage{academicons}
\usepackage{cancel}
\usepackage[ruled,vlined]{algorithm2e}
\SetKw{KwBy}{by}
\SetKw{KwTo}{in}
\SetKwComment{Comment}{/* }{ */}

\usepackage{tikz,xcolor,hyperref}

\definecolor{lime}{HTML}{A6CE39}
\DeclareRobustCommand{\orcidicon}{
	\begin{tikzpicture}
	\draw[lime, fill=lime] (0,0) 
	circle [radius=0.16] 
	node[white] {{\fontfamily{qag}\selectfont \tiny ID}};
	\draw[white, fill=white] (-0.0625,0.095) 
	circle [radius=0.007];
	\end{tikzpicture}
	\hspace{-2mm}
}
\foreach \x in {A, ..., Z}{
	\expandafter\xdef\csname orcid\x\endcsname{\noexpand\href{https://orcid.org/\csname orcidauthor\x\endcsname}{\noexpand\orcidicon}}
}

\protect

\newcommand{\orcid}[1]{\href{https://orcid.org/#1}{\includegraphics[width=8pt]{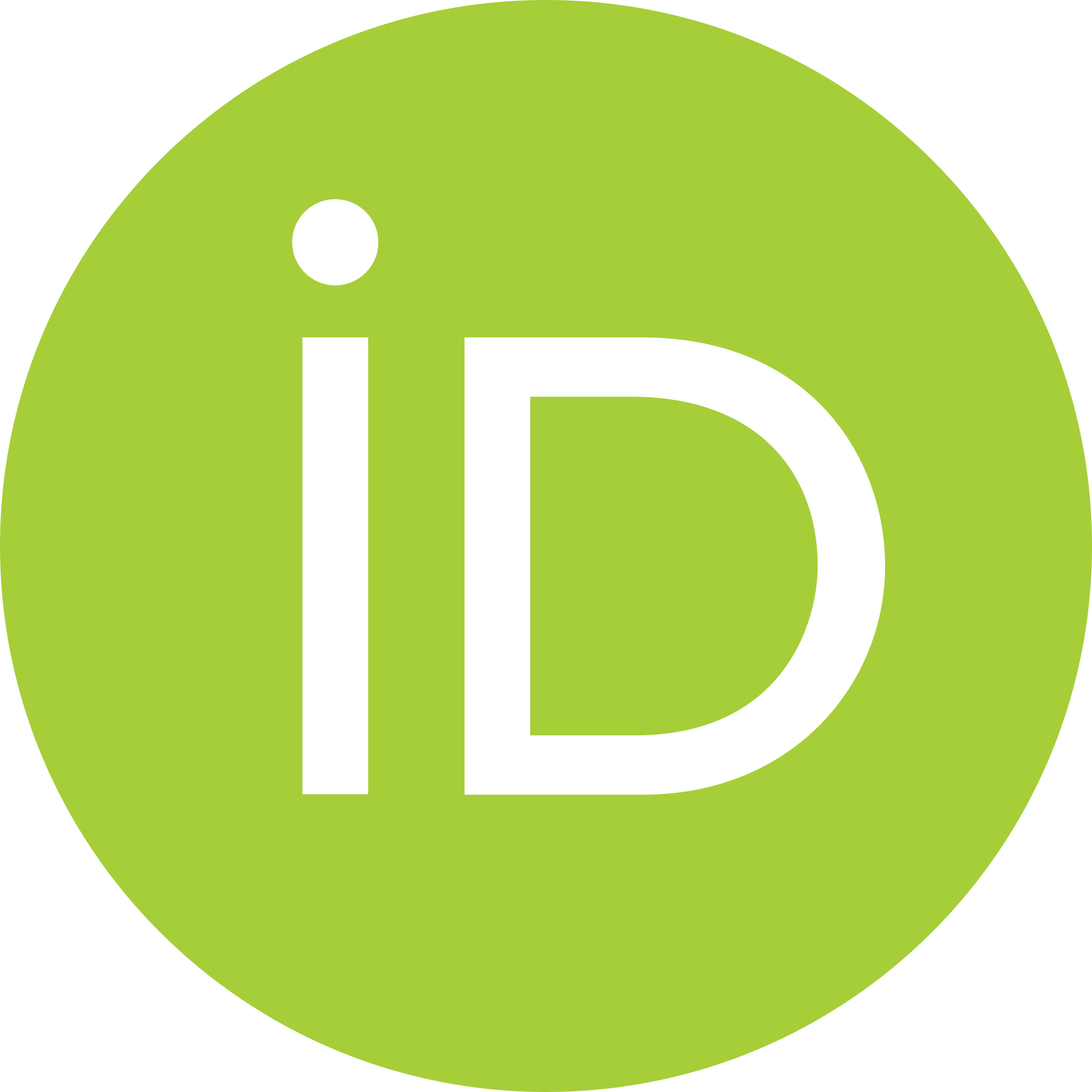}}}
\begin{document}
\title{A speckle filter for Sentinel-1 SAR\\ Ground Range Detected data based on\\ Residual Convolutional Neural Networks}

\author{\IEEEauthorblockN{Alessandro Sebastianelli\orcidA{}, \IEEEmembership{Member, IEEE},
Maria Pia Del Rosso\orcidB{}, \IEEEmembership{Member, IEEE},\\
Silvia L. Ullo\orcidC{}, \IEEEmembership{Senior Member, IEEE}, 
Paolo Gamba\orcidD{}, \IEEEmembership{Fellow, IEEE} 
}
\thanks{Alessandro Sebastianelli, Maria Pia Del Rosso, Silvia Liberata Ullo are with the Departement of Engineering, University of Sannio, 82100 Benevento, Italy (email: sebastianelli@unisannio.it;  mpdelrosso@unisannio.it;  ullo@unisannio.it).\\ Paolo Gamba is with the Department of Electrical, Computer and Biomedical Engineering, University of Pavia, 27100 Pavia, Italy (e-mail:
paolo.gamba@unipv.it)}}

\maketitle

\begin{abstract}
In recent years, machine learning (ML) algorithms have become widespread in all the fields of remote sensing (RS) and earth observation (EO). This has allowed the rapid development of new procedures to solve problems affecting these sectors. In this context, this work aims at presenting a novel method for filtering speckle noise from Sentinel-1 ground range detected (GRD) data by applying deep learning (DL) algorithms, based on convolutional neural networks (CNNs). The paper provides an easy yet very effective approach to extract the large amount of training data needed for DL approaches in this challenging case. The experimental results on simulated speckled images and an actual SAR dataset show a clear improvement with respect to the state of the art in terms of peak signal-to-noise ratio (PSNR), structural similarity index (SSIM), equivalent number of looks (ENL), proving the effectiveness of the proposed architecture. 
\end{abstract}

\IEEEdisplaynontitleabstractindextext
\IEEEpeerreviewmaketitle

\section{Introduction}

\begin{figure}[!ht]
     \centering
     
     \resizebox{\columnwidth}{!}{\begin{tabular}{cc}
     \includegraphics[width=\columnwidth]{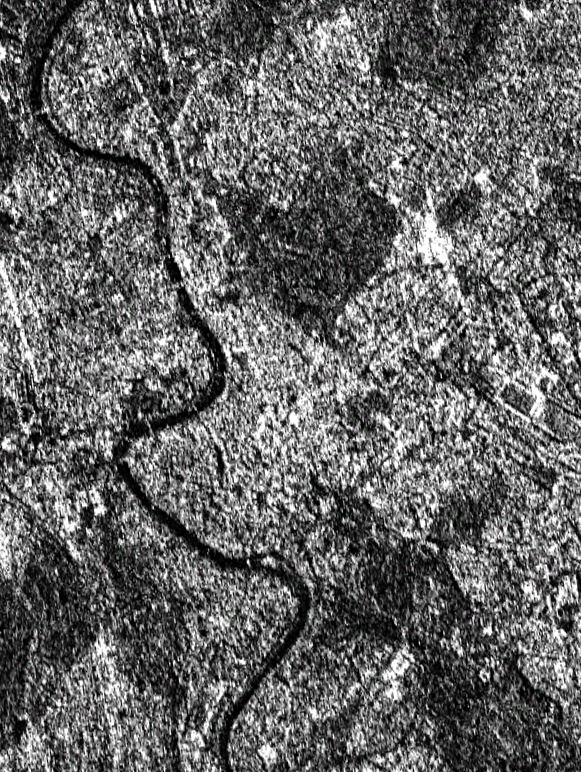} &
     \includegraphics[width=\columnwidth]{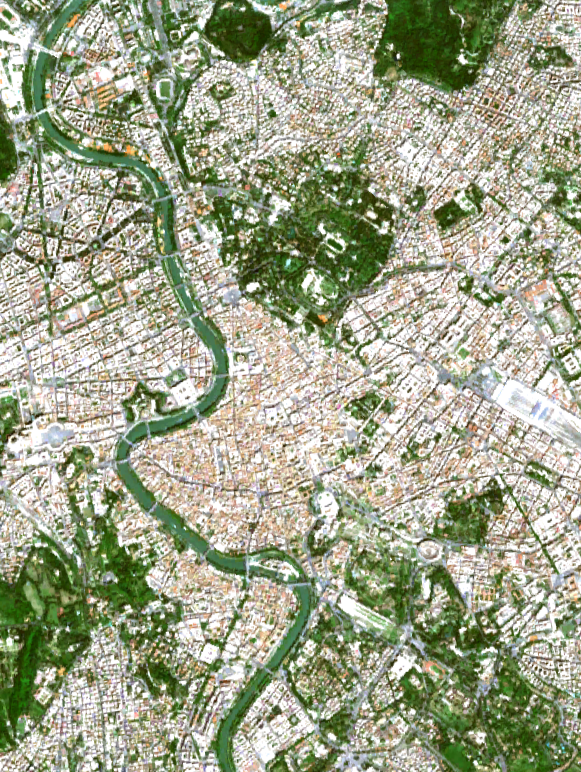} \\
     \end{tabular}}
     \caption{Comparison between Sentinel-1 (on the left) and Sentinel-2 (on the right) acquisitions on 2022-03-17 and 2022-04-30 respectively, focusing on Rome (Italy): a  Single Look Complex (SLC) Interferometric Wide Swath (IW) mode  SAR image in VV polarization versus  a Level2 (L2)-A optical image in true colors (composition of RGB channels).}
     \label{fig:s1_s2_sample}
\end{figure}

Remote Sensing (RS) data are increasingly used in many research fields to extract a large amount of different information. Their variety, ranging from multispectral and hyperspectral to active radar data, allows to select the most appropriate type   for each application of interest. However, most of the times, researchers have to deal with several sources of noise and disturbances corrupting the quality of the data and impacting the correctness of final results, and it becomes crucial to understand the best way to tackle them in order to recover data integrity.

In this regard, an interesting study has been carried out in \cite{4072455}, where the authors have  studied how different sources of  external and internal noise, atmospheric effects, quantization noise, sensor detector/preamplifier processes,  influence the accuracy  of the multispectral classification in RS systems. 

In the years, many methods have been developed to reduce noise effects on RS images, although these methods are dependent on the type of data and the source of noise, and may have an impact on the useful data while removing the noise. Thankfully, the most recent techniques have achieved better noise removing performances while preserving useful information.

For example, Lv ZhiYong et al. in \cite{article} and  \cite{article3} have introduced a filtering technique for Very High Resolution (VHR) RS images, the multi-scale filter profile (MFP)-based framework (MFPF), which improves  the classification results based on different shapes and scales of the objects, while reducing the noise.

Optical and Synthetic Aperture Radar (SAR) RS images are usually affected in a different way because of the intrinsic difference of the transmitted and received signals. In particular, optical RS images are usually contaminated by Gaussian white noise, and  the interference between different features is stronger especially in the heterogeneous regions, where it can not be removed easily \cite{2017ZhuBo}. Moreover, cloud cover represents a serious limitation in the use of optical RS data \cite{PRUDENTE2020100414, Asner2001}, resulting in a noise-like disturbance.

SAR data are also affected by several types of noise, but the speckle is of high relevance, resulting into a  constructive- or destructive-interference phenomenon occurring in each resolution cell,  due to its larger dimension with respect to the wavelength of the incident electromagnetic field \cite{6352296, 9690877}. The speckle  confers  a granular aspect to the final image by spoiling its quality and interpretation \cite{goodman1976some}, \cite{zhang2021robust, liu2020training}, unlike optical data, that when not affected by cloud cover are characterized by clear and recognizable features, as evident in Figure \ref{fig:s1_s2_sample}.

The presence of the speckle noise influences the results of many applications based on SAR data. Even techniques making use of interferometric data for the estimation of terrain displacements \cite{chen2013comparative} suffer from the noisy data and, as a consequence,  filtering methods (e.g. Multilooking) need to be introduced in the interferometric processing chain. Speckle also affects the phase of scattering coefficients for the analyzed surface and degrades the   polarimetric information of the signal \cite{lopez2003polarimetric, maitra2013characterization}.

In recent years, the use of SAR data for RS applications is hugely increased, and especially the employment of the ground range detected (GRD) products, mainly for two reasons: 1) the lower computational load with respect to other SAR products; 2) their availability on cloud platforms like the Google Earth Engine (GEE). In literature, many works give important examples of how GRD SAR data can help in offering solutions to practical cases of interest. In \cite{GRD_case1}, a new methodology for rapid flood mapping is presented making use of Sentinel-1 GRD images, and the authors highlight "the potentiality of the proposed methodology,  particularly oriented toward the end-user community". In \cite{Gamba4, Gamba5}, authors describe an urban change analysis based on Sentinel-1 GRD data. They underline how to handle big data, intensive pre-processing steps and computations are run in GEE. Some other studies exploit the complete polarimetric information contained in dual-pol GRD SAR data, useful for crop growth monitoring \cite{GRD_case2}, \cite{GRD_case3}. The first of these two has presented an innovative clustering framework working to facilitate the operational use of Sentinel-1 GRD SAR data for agricultural applications. The second work focuses on the Speckle Noise, Single Look Complex (SLC) and GRD data, for urban features identification.

In general, what appears interesting is the increase of SAR data employment in those application fields where for years optical data have dominated, such as vegetation monitoring and feature extraction from urban areas, and the complimentary use of SAR and optical data opening up new applications \cite{rs13173535, rs70201320, xu2022dynamic}.

For all above specified, in this article the authors have decided to focus their work on speckle filtering in GRD SAR data. Specifically, this article involves the use of  a DL model based on residual CNNs to filter speckle noise in GRD SAR data, and it presents  many novelties with respect to the-state-of-the-art, that will be explained in detail ahead in this paper, and are here summarized as follows: {$1)$ a new method for generating the dataset, adding the simulated speckle to the generated synthetic ground truth, and without applying a log-transformation to the data;} $2)$  the use of a residual CNN with a subtractive layer, trained on the designed dataset, namely the non-simulated speckle-free SAR-like data and simulated speckled SAR data; $3)$ the introduction of a learning rate scheduler combined with a multi-term loss function inside the CNN workflow for achieving a better training and for producing smoother results; $4)$ the use of a long time-series of GRD SAR multi-looked intensity data   to build the synthetic (SAR-like) ground-truth as a speckle-free representation of the input; 
$5)$ experiments also on non-SAR data when Gaussian noise is applied, besides the speckle one, and with different power levels, to verify the model performance and transferability.

Comparison and discussion of results on the built dataset but also on real data (these latter obtained in the right format through a detailed description of necessary steps on SNAP) have been conducted, by demonstrating how the proposed model  achieves  comparable or higher performances than others   based on similar (subtractive) methods or different methods such as for instance the logarithmic-state space transform or the divisional layers. All that will be extensively presented and discussed in the Results section.

It is worth to highlight that both the dataset and the model have been developed from scratch and made available on GitHub, with the code open-source, for further analysis and investigation \cite{code, dataset}. The authors want also to highlight that to acknowledge Prof. Landgrebe inspiring research instrumental in the inception and growth of spectral remote sensing image analysis and processing, they made all the best to add explanations in such a way  that the code could be easily usable by  interest researchers, and the philosophy to promote a research that could work in the direction of remote sensing image analysis growth has always pushed us to prefer open data (i.e. Sentinel data from ESA Copernicus mission), free tools, open cloud repository and processing platform (i.e. GEE).

The remaining part of the article is structured as follows: Section II presents the background on the types of speckle filtering methods found in the literature, and in Section III, the dataset generated for training the neural network is explained, with the description of the proposed method. In Section IV results and discussion are given. Lastly, conclusions end the manuscript.

\section{Background}
As outlined before, the problem to clean SAR images from the speckle noise has engaged researchers for years, and several speckle filtering methods have been proposed. \\
Among different ways of classifying speckle filters, the categorization chosen in this article refers to three macro categories: $1)$ single-product speckle filtering, $2)$ multi-temporal speckle filtering and $3)$ AI-based speckle filtering.

\textbf{Single-product speckle filtering} methods are mainly based on  mathematical models of the speckle phenomenon, and a usual approach in this category is the spatial multilooking, that can reduce the speckle by means of  incoherent averaging of many sub-aperture acquisitions (looks). This process normally includes the sub-sampling of data and the ground-projection, by leading to the increase of the radiometric resolution but to the decrease of the spatial one. Another disadvantage in this case is that the phase information is lost \cite{oliver2004understanding}. In the single-product speckle filtering category, another typical approach involves the use of adaptive and non-adaptive filters. Adaptive filters are better at preserving edges and details in high-texture areas (e.g. urban areas), and some examples are given by the Lee, the Frost, and the Refined Gamma Maximum-A-Posteriori (RGMAP) filters \cite{banerjee2021survey, baraldi1995refined, lee1986speckle}. With respect to the non-adaptive filters, there are two forms of speckle filtering: one based on the mean and one based upon the median. They are simpler to implement and require less computational power. However, they can degrade the acquisition and result into an information loss \cite{frery1993non, parihar2015radar}.

Gomez et al. in \cite{Gomez1} propose the quantitative evaluation of the most common single-product despeckling filters, by introducing also a new image-quality index. 
This latter provides consistent results with commonly used quality measures and ranks the filters also in agreement with their visual analysis. Another interesting framework for benchmarking despleckling filters of single SAR images is presented in \cite{Riccio2}, where  a quantitative  evaluation of the their despeckling capabilities is  given with respect to aspects such as edge smoothing, texture
and target feature preservation, under five designed scenes.

\textbf{Multi-temporal speckle filtering} includes the second category of speckle filtering techniques applied on a dataset of co-registered acquisitions. This approach may allow achieving very high signal-to-noise ratios by preserving the spatial resolution, as presented in \cite{DiMartino28}. 
 A two-step procedure is presented in \cite{Quegan2001}, which under the hypothesis of neglected correlation among the bands applies first an intensity averaging of the data, and then an additional averaging in the spatial domain, with  unavoidable losses of spatial resolution for the filtered products. A two-step approach is also introduced in \cite{DiMartino30},yet in this case the first step performs a temporal multilooking on stable pixels selected through a Kullback–Leibler procedure In \cite{DiMartino31} temporal and spatial filtering is exploited as well, yet the correlation among bands is not neglected, and the overall procedure does not reduce the spatial resolution.
 In \cite{DiMartino33}, instead of processing the whole time series, a superimage is created by collecting all the images over time (similarly to what done in \cite{DiMartino30}) and a ratio image  between the corresponding noisy image and the superimage is formed. The despleckled image is obtained by multiplying the superimage by the ratio image clean through a RuLoG denoiser. An interesting benchmarking framework has been presented in \cite{Riccio1}, built over the  hypothesis of temporally uncorrelated bands, to evaluate  different  methods for multitemporal  despeckling. Based on the proposed framework, their resulting performance  is well in agreement with (qualitative) visual inspections done by SAR specialists, by proving the effectiveness of such a tool.\\
\textbf{AI-based speckle filtering} In recent years, the enormous spread  of Artificial Intelligence (AI)-based methods for EO has given rise to the development of new speckle filtering techniques based on the use of Deep Learning (DL) algorithms, and in particular of CNNs, as demonstrated by the large number of works found in the literature.

Chierchia et al. \cite{8128234} have been the pioneers in introducing methods for speckle denoising based on Convolutional Neural Networks. They adapted the method of Gaussian noise removal based on CNNs to the speckle problem. They took as baseline the residual approach, such as not recovering the filtered image but estimating the speckle noise and subtracting it directly to the noisy image. Since the speckle is a multiplicative type of noise, they transformed the image values in log scale in order to apply directly the subtraction operation and then invert the transformation to obtain the filtered image. Their model is based on Convolutional, Batch Normalization and ReLu blocks.

Wang et al. \cite{Wang2017}, on the contrary, use a division residual method, estimating directly the speckle from the input images and introducing a componentwise division residual layer into the network. Their proposed Image Despeckling Convolutional Neural Network (ID-CNN) consists of several convolutional layers along with batch normalization and rectified linear unit (ReLU) activation function. In \cite{wang2018generating}, they improved the ID-CNN to create despeckled SAR images to be used as inputs to a generative architecture for colorizing them and  generating optical-like images.

Lattari et al. \cite{lattari2019deep} propose a U-Net convolutional neural network, modified and adapted for the despeckling task. The main novelty of their approach was the possibility to preserve edges, and permanent scatter points while producing smooth solutions in homogeneous regions. This improvement allowed to obtain high-quality filtered images with no additional artefacts.

In \cite{cozzolino2020nonlocal}, Cozzolino et al. propose a simple CNN-powered nonlocal means filter, that is, plain pixel-wise nonlocal means in which the filter weights are computed by means of a dedicated convolutional network.

Dalsasso et al. \cite{dalsasso1} handle the speckling problem by applying two different strategies: first, they apply directly  a CNN model trained to remove additive white Gaussian noise from natural images, to a recently proposed SAR speckle removal framework: MuLoG (MUlti-channel LOgarithm with Gaussian denoising), without performing a training on SAR images, then they train the CNN used to remove additive white Gaussian noise on speckle-free SAR images. In \cite{dalsasso2021sar2sar}, they adapt the noise2noise framework to SAR despeckling, based on a compensation of temporal changes and
a loss function adapted to the statistics of speckle. In particular, the training of the network is performed in three steps: A) first on images with synthetically generated speckle; B) then on pairs of images extracted randomly from a timeseries, where the second image is compensated for changes based on reflectivities estimated with the network trained in (A); C) finally a refinement step is performed where the network weights in (B) are used to obtain a better compensation for changes.

In \cite{molini2021speckle2void}, Molini et al., inspired by recent works on blind-spot denoising networks, propose a self-supervised Bayesian despeckling model, trained employing only noisy SAR images.

Other useful information, used for positioning our method in the current state of the art for AI-based speckle filters, is reported in Table \ref{tab:categorization}, showing data used for training, the SAR product, if data log-scaling is applied or not, and an identification of the proposed strategy.

\begin{table}[!ht]
    \centering
    \resizebox{1\columnwidth}{!}{
    \begin{tabular}{lllll}
        \toprule
        Paper & Data & SAR Products & Log-scale & Strategy\\
        \midrule
        \cite{8128234} & COSMO-SkyMED & SLC & yes & subtractive\\
        \cite{8128234} & COSMO-SkyMED & SLC & yes & subtractive\\
        
        \cite{Wang2017} & - & - & no & divisional\\
        \cite{wang2018generating} & RADARSAT-1 & SLC & no & divisional\\
        \cite{lattari2019deep} & Sentinel-1 & SLC & yes & subtractive\\
         & COSMO-SkyMED &  &  & \\
        \cite{cozzolino2020nonlocal} & RADARSAT & SLC & yes & non-local CNN\\
         & TerraSAR-X &  &  & \\
         & COSMO-SkyMED &  &  & \\
        \cite{dalsasso1} & TerraSAR-X & SLC & yes & subtractive\\
         & Sentinel-1 &  &  & \\
        \cite{dalsasso2021sar2sar} & Sentinel-1 & SLC & yes & subtractive\\
        \cite{molini2021speckle2void} & TerraSAR-X & SLC & no & Blind-spot CNN\\
        \midrule
        Proposed & Sentinel-1 & GRD & no & subtractive\\
        \bottomrule
    \end{tabular}}
    \caption{Categorization of AI-based Speckle Filters}
    \label{tab:categorization}
\end{table}

\section{Data and Methods}
In the following sections, both the created dataset and the developed neural network architecture will be presented and explained in details. 

\subsection{Dataset creation for the proposed CNN} \label{sec:dataset}
Since the proposed method for filtering the SAR speckle is based on CNNs, trained in a supervised way, both the input and the expected output (ground truth) should be available for training the model.

While in some other works found in the literature the models are trained in a supervised way by using optical images as reference and by fine-tuning the weights with SAR data \cite{Wang2017, lattari2019deep}, the approach here proposed is based on the generation of a dataset by starting from real SAR data, leading to "SAR-like" reference images. It is important  to highlight the "SAR-like" aspect, indeed the dataset is a realistic representation of the truth, since  the initial real data have been averaged, and the speckle simulated, as explained ahead. Therefore, the generated ground truth is referred to as a synthetic one.

\begin{figure}[!ht]
    \centering
    \includegraphics[width=0.96\columnwidth]{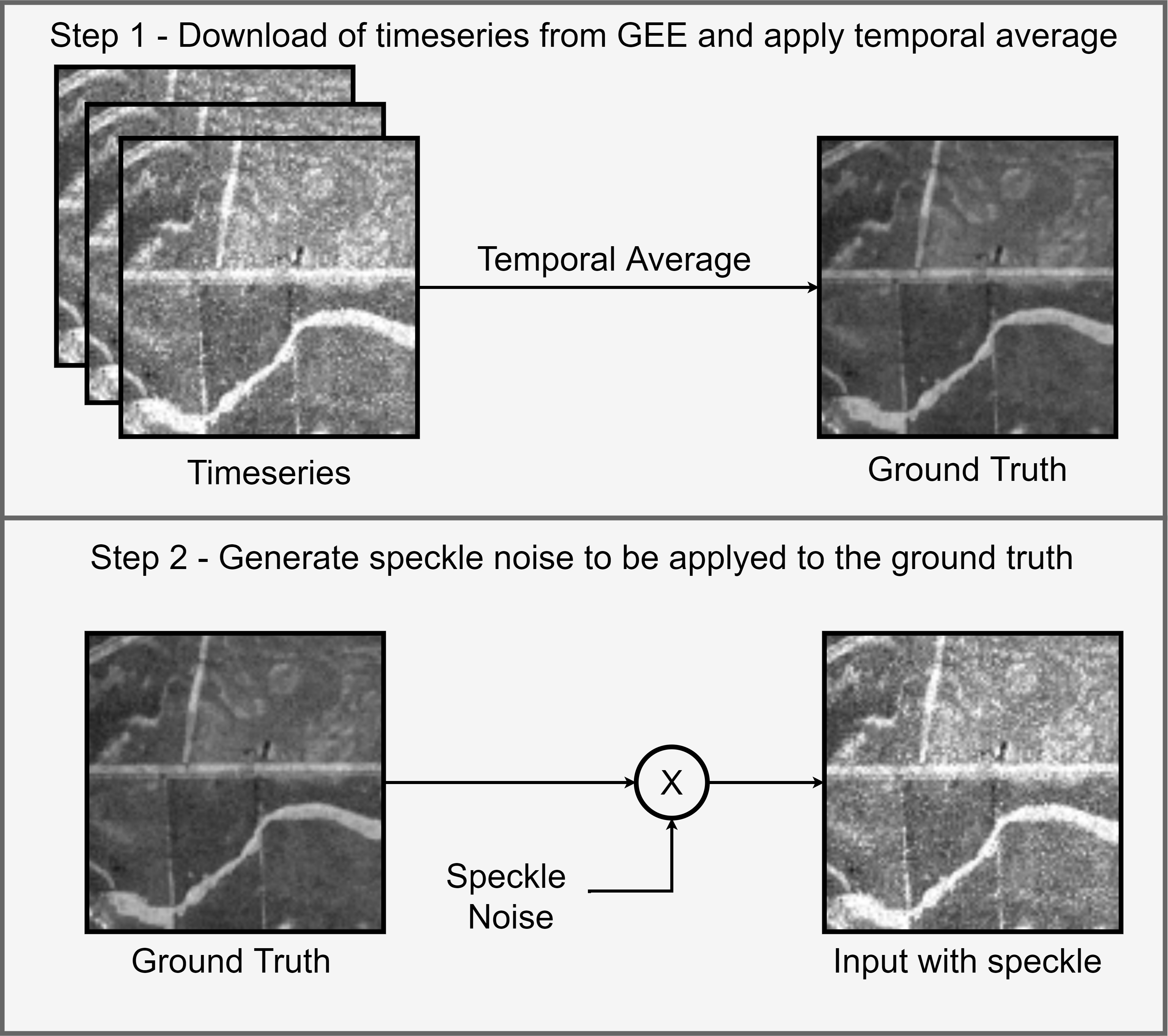}
    \caption{Dataset creation pipeline. On top: step 1 - Download of SAR time-series from GEE and application of the temporal average to get the ground truth. On bottom: step 2 - generation of speckle noise with Gamma distribution to be multiplied with the ground truth to get the noisy input (step 3).}
    \label{fig:dataset}
\end{figure}

Differently from \cite{dalsasso1}, a long time-series of GRD Sentinel-1 images are downloaded from GEE and temporally-averaged to get the ground truth. The second step consists in generating the speckle noise with the most representative statistical distribution, and lastly, the generated speckle is multiplied with the ground truth. This process is summarized in Figure \ref{fig:dataset}, where the  main steps for creating the dataset are sketched.\\
At this point, it is important to highlight that the GRD Sentinel-\nobreak 1 images have been downloaded from GEE, but the results do not change if they are directly taken from the Copernicus Open Acess Hub, as demonstrated through the analysis shown ahead in the Section IV.C where also the SNAP processing chain is presented.

\textbf{Generation of Synthetic Ground Truth.}   Let \textbf{X}$_{lat,lon}\in \mathbb{R}^{W\times H \times P}$ be a single Sentinel-1 SAR acquisition, with a width $W$, a height $H$ and a specific polarization $P$,  acquired in the geographical region defined by latitude $lat$ and longitude $lon$. Based on the tool proposed in  ~\cite{sebastianelli2020automatic}, a long time-series of Sentinel-1 acquisitions $\Vec{\mathbf{X}}_{lat,lon}\in \mathbb{R}^{T\times W\times H \times P}$ has been collected, with $T$ denoting its length (note: the entire collection has been used since Sentinel-1 mission started). 

Data have been downloaded from the Google Earth Engine (GEE) catalog ~\cite{gorelick2017google}, containing the full Sentinel-1 archive of GRD products. GRD data consist of focused SAR data that have been detected, multi-looked and projected to ground range using an Earth ellipsoid model. Phase information is lost. The pixels for this product are approximately square-shaped, with reduced speckle at the cost of a lower spatial resolution \cite{grd}.

Interferometric Wide (IW) swath GRD High Resolution Sentinel-1 intensities have been used to create the dataset by employing the GEE collection with raw power values, COPERNICUS/S1\_GRD\_FLOAT, not log-scaled \cite{mullissa2021sentinel, geee1, geee2}, indeed by applying a homomorphic processing, in which the multiplicative noise is transformed into an additive noise by taking the logarithm of the observed data, it could result in failing to preserve sharp features (e.g. edges) and introduce artifacts in the denoised image \cite{patel2013separated, Wang2017}.

To obtain the ground truth from Sentinel-1 acquisitions with a speckle influence as lower as possible,  a temporal average of intensity data $\Vec{\mathbf{X}}_{lat,lon}$  has been carried out, as expressed by equation \eqref{eqn:temporal_average}. The solution of averaging the intensities has been adopted since averaging the amplitudes would lead to a biased result and leave more residual speckle fluctuations than those observed when averaging the intensities. The rational behind this choice is based on Goodman's speckle model, from which one can say that the maximum likelihood estimator of the underlying reflectivity is obtained by averaging the SAR intensities.
\begin{equation}
    \centering
    \mathbf{X}_{lat,lon}^* = \frac{1}{T}\sum^T  \Vec{\mathbf{X}}_{lat,lon}\ ,\ \ \  \mathbf{X}_{lat,lon}^* \in \mathbb{R}^{W\times H \times P}
    \label{eqn:temporal_average}
\end{equation}

The final input instead has been obtained as anticipated before by multiplying the ground truth $\mathbf{X}_{lat,lon}^*$ by the speckle $\mathbf{S}$, generated through the statistical model presented in the following \textbf{Speckle generation} paragraph.

\textbf{Speckle generation} To understand how the final dataset has been generated, a short description of the traditional Speckle model is given.

One of the most common used models for SAR speckle is the multiplicative noise model \cite{goodman1976some} expressed by the equation:

\begin{equation}
    \centering
    Y = X\cdot S
\end{equation}

where $Y \in \mathbb{R}^{W\times H}$ is the observed intensity acquisition, $X \in \mathbb{R}^{W\times H}$ is the noise-free version of the acquisition and $S \in \mathbb{R}^{W\times H}$ is the speckle noise.  By considering a multi-looking (L looks) SAR acquisition, $S$ follows the Gamma distribution with unitary mean and variance $1/L$.

Since we used IW GRD High Resolution Sentinel-1 intensities data, downloaded from the GEE repository,  in accordance with what is specified in \cite{s1grdlooks}, the speckle has been generated by taking samples from the Gamma distribution with a number L of looks equals to  4.  The speckle must have the same dimensions of the ground truth, thus $\mathbf{S} \in \mathbb{R}^{W\times H \times P}$. More details about the statistical models for the speckle can be found in Appendix \ref{app:speckle}.  
\\

\textbf{Generation of SAR-like images}. The input image is then derived using the equation:
\begin{equation}
    \centering
    \mathbf{X}_{lat,lon}^s = \mathbf{X}_{lat,lon}^* \times \mathbf{S}\ ,\ \ \  \mathbf{X}_{lat,lon}^s \in \mathbb{R}^{W\times H \times P}
    \label{eqn:input_generation}
\end{equation}

This process has been repeated for an arbitrary number of times (this choice will define the size of the final dataset) by randomly selecting latitude $lat$ and longitude $lon$ values, distributed over the land surface.

\textbf{Dataset composition}. The inputs and the ground truths are then grouped in vectors, in order to form the dataset, as expressed through equations (\ref{eqn:vectors}a) and (\ref{eqn:vectors}b).

\begin{subequations}
    \begin{align}
        \Vec{\mathbf{X}}_{inputs} &= \left[\mathbf{X}_{lat_0,lon_0}^s,\dots,\mathbf{X}_{lat_N,lon_N}^s \right]\\
        \Vec{\mathbf{X}}_{ground\_ truths} &= \left[\mathbf{X}_{lat_0,lon_0}^*,\dots,\mathbf{X}_{lat_N,lon_N}^* \right]
    \end{align}
    \label{eqn:vectors}
\end{subequations}

where $\Vec{\mathbf{X}}_{inputs}, \Vec{\mathbf{X}}_{ground\_ truths} \in \mathbb{R}^{N\times W \times H \times P}$, with $N$ representing the number of Sentinel-1 acquisitions on the Earth surface.

The dataset is then  divided into three sub-datasets, respectively the training dataset $\Vec{\mathbf{X}}_{inputs}^t, \Vec{\mathbf{X}}_{ground\_ truths}^t \in \mathbb{R}^{M\times W \times H \times P}$,  the validation dataset $\Vec{\mathbf{X}}_{inputs}^v, \Vec{\mathbf{X}}_{ground\_ truths}^v \in \mathbb{R}^{Q\times W \times H \times P}$ and  the testing dataset $\Vec{\mathbf{X}}_{inputs}^g, \Vec{\mathbf{X}}_{ground\_ truths}^g \in \mathbb{R}^{R\times W \times H \times P}$, where $M=75.8 \%$ of $N$, $Q=22.8 \%$ of $N$ and $R=1.4 \%$ of $N$.

The final dataset contains $2637$ Sentinel-1 acquisitions without speckle and $2637$ corresponding acquisitions with speckle. Using the training-validation-test split factors, respectively of $75.8 \%$, $22.8 \%$ and $1.4 \%$ as above specified, the resulting training dataset is composed of $2000$ samples, the validation dataset of $600$ samples and the testing dataset of $37$ samples. Data augmentation has also been  applied to these datasets, for artificially increasing their original size, through transformations of the starting images (e.g. rotation, crops, etc.). A version of the proposed dataset has been made available online  open access \cite{dataset}.

\subsection{Proposed approach}

\begin{figure*}[!ht]
    \centering
    \includegraphics[width=2\columnwidth]{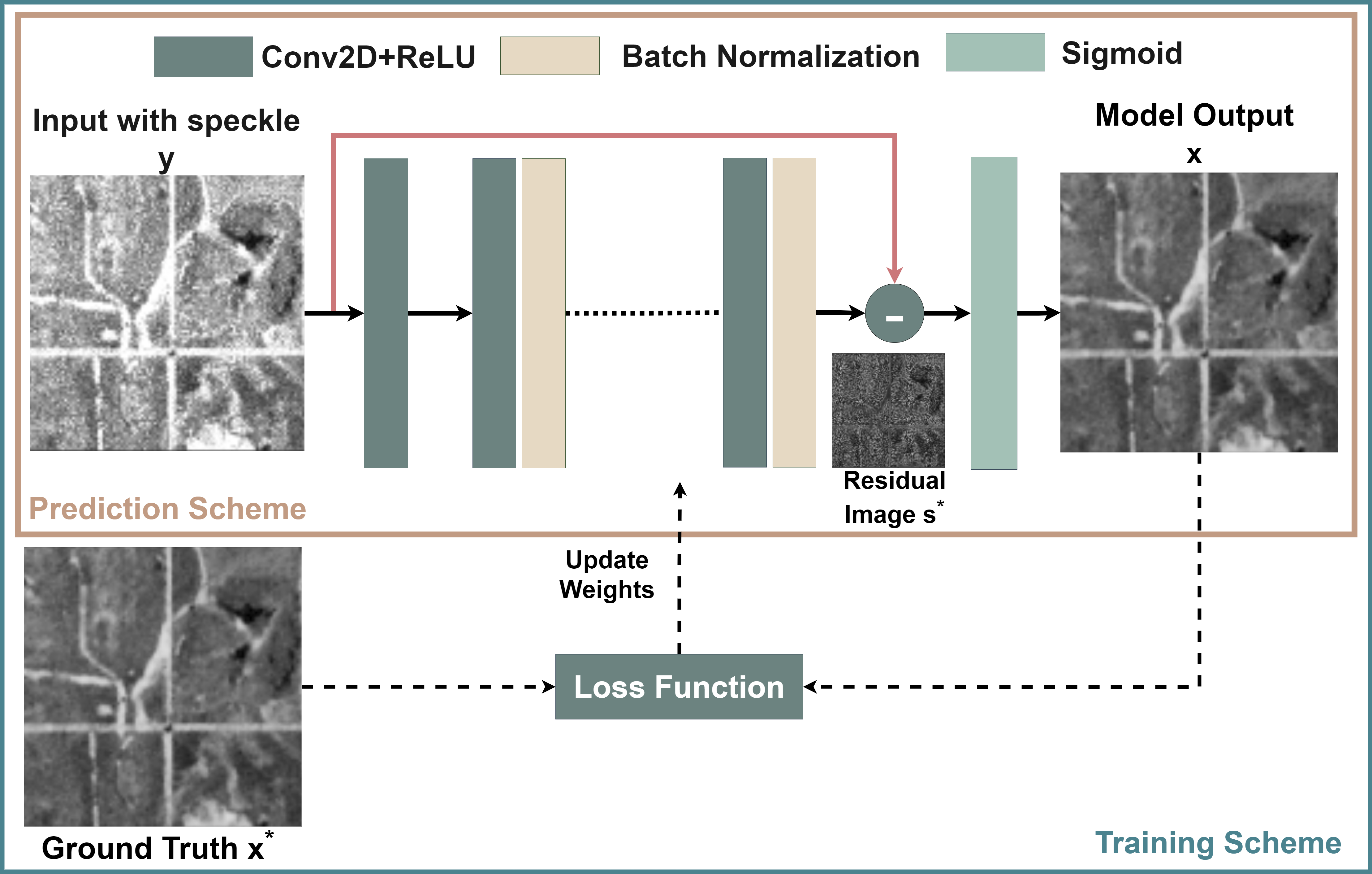}
    \caption{Proposed model for speckle filtering.}
    \label{fig:model}
\end{figure*}

As reported in Table \ref{tab:categorization}, no other works in the literature show a similar approach, with a subtractive layer in a residual CNN with GEE GRD not log-scaled data.

Moreover, although the starting point for realizing the proposed network refers to speckled SAR data, as presented in the previous paragraph by bringing to the creation of the testing dataset, the further intent of this work has been to realize a  model that should not take into consideration the nature of the noise affecting the input image, in such a way  to create a DL architecture capable of picking up any type of noise and eliminating it through a subtractive layer. As shown  in the results section,  the proposed model has been tested with different sources of noise, at different power and with images of different nature, by leading to interesting outcomes and observations.

In particular, the proposed approach is based on the extraction of the speckle from the input Sentinel-1 acquisition through a CNN. The extracted speckle will be then subtracted from the input image through a skip connection, that makes the model also residual, and a subtraction layer \cite{book}. The output of the subtraction layer will be the filtered Sentinel-1 acquisition. The model is essentially composed of three components: $1)$ the CNN, $2)$ the skip connection and $3)$ the subtraction layer, as shown in Figure \ref{fig:model}.

The residual CNN is composed of repeated fundamental blocks, such as 2D Convolutions, ReLUs and Batch Normalizations. It mainly deals with the extraction of speckle noise from the input data. The skip connection, the red arrow in Figure \ref{fig:model}, which connects the input with the subtraction layer, is essentially a connection used to skip the CNN. The subtraction layer applies the mathematical subtraction between the input data, through the skip connection and the CNN output. The result, after the last Sigmoid layer, is the filtered Sentinel-1 acquisition.

During the training of the model, lasting 50 epochs, a batch of 16 Sentinel-1 acquisitions from the training dataset $\Vec{\mathbf{X}}_{inputs}^t$ is used as feedforward input for the model. The predicted images \textbf{Y}$\in\mathbb{R}^{W\times H \times P}$ are compared with the corresponding ground-truth acquisitions from the training dataset $\Vec{\mathbf{X}}_{ground\_ truths}^t$, through the loss function, whose description is given ahead. Using this loss function and its gradient, the Adam optimizer ~\cite{zhang2018improved} updates the weights of the network. 
The Learning Rate (LR) is initially set to $0.002$ and then updated through a LR Scheduler implemented ad hoc. The scheduler updates the LR every 5 epochs following the rule:

\begin{equation}
    \centering
    learning\ rate^{(i)} = learning\ rate^{(i-5)} * decay\ rate
\end{equation}

where $i$ indicates the epoch and \textit{decay rate} is set to $0.8$. The total number of parameters to be optimized is $447.425$.

Contrarily to the techniques that transform the multiplicative model of the speckle into an additive one, through a logarithmic transformation, or to the techniques that use divisional layers, as anticipated, the proposed method adopts a residual strategy based on subtraction layers \cite{lattari2019deep}, done  without log-scaling the data to preserve their statistical characteristics (for this reason GEE data in the FLOAT format have been used) and by avoiding the use of divisional layer that may lead to computational error (e.g. the division by zero could lead to an inadvertent stop of the training). With this configuration, the residual noise is expressed as difference between the noisy and the clean image as expressed by the equation:

\begin{equation}
    \centering
    s^* = y - x
    \label{eqn:difference}
\end{equation}

where $s^*$ denotes the residual image, $x$ denotes the clean image and $y$ the image affected by speckle. The model is  trained to find the function:

\begin{equation}
    \centering
    f(y,\vec{\omega}) = x
\end{equation}

where $\vec{\omega}$ are the weights of the model. Through the optimization of $\vec{\omega}$, by minimizing a customized loss function $L$:

\begin{equation}
    \centering
    \begin{split}
    L&(x ,x^*) = \alpha  \frac{1}{WH} \sum^W_{w=1}\sum_{h=1}^H [x_{w,h} - x^*_{w,h}]^2 + \\
     &+ \beta \frac{(2\mu_{x}\mu_{x^*}+c_1)(2\sigma_{xx^*}+c_2)}{(\mu^2_{x}+\mu^2_{x^*}+c_1)(\sigma^2_{x}+\sigma^2_{x^*}+c_2)} +\\
     &+ \gamma \sum_{w=1}^{W}\sum_{h=1}^H \sqrt{(x_{w+1,h}-x_{w,h})^2 + (x_{w,h+1}-x_{w,h})^2}
    \end{split}
    \label{eqn:loss}
\end{equation}

the model learns as well as possible the real map between $y$ and $x$. In Equation \ref{eqn:loss}, $x^*$ is the ground truth, $x$ is the model output;  the first term, that is the square of the Euclidean distance, is responsible for the minimization of the distance of the pixels' values between the model output and the ground truth, the second term is responsible for  the preservation of the structural similarity between the model output and the ground truth. Since the Euclidean loss has shown to work well on many image restoration problems, but it often results in artifacts on the filtered image, a third term, the  Total Variation Loss has been included,   to force the model to create smoother results.

The parameters $\alpha$ and $\beta$ are initially set to $1$, meaning that the first two terms of the loss have the same importance during the minimization process, while the parameter $\gamma$ has been fined tuned thorough several tests and it has been set to $0.00001$ in order to introduce the right amount of smoothness in the predicted image.

In the second term of $L$,  $\mu_{\hat{s}}$ is the average of $\hat{s}$, $\mu_{s^*}$ is the average of $s^*$, $\sigma_{\hat{s}}^2$ is the variance of $\hat{s}$, $\sigma_{s^*}^2$ is the variance of $s^*$, $\sigma_{\hat{s}s^*}^2$ is the covariance of $\hat{s}$ and  $s^*$, $c_1 = (k_1D)^2$ and $c_2 = (k_2D)^2$ are two variables that stabilize the division with a weak denominator, $D$ is the dynamic range of the pixel-values (typical $2^{\#BitsPerPixel}-1$), $k_1=0.01$ and $k_2=0.03$ by default.

During the inference, the despeckled image $x$ will be obtained from the original noisy image $y$ by subtracting the residual noise $s^*$ predicted by the CNN in accordance with the relation described by equation \ref{eqn:difference}.

Some pre-processing steps are necessary before the training:  after removing the outliers by clipping the values to the 90th percentile, both the input and the expected output are  normalized into the range $[0,1]$,  by using the min-max normalization as given by the formula:
\begin{equation}
    \centering
    \mathbf{Y} = \frac{\mathbf{X} - min}{(max - min)+\varepsilon}
    \label{eqn:min-max}
\end{equation}
Data normalization is a very common technique in ML and it allows the model to learn faster and to produce better results \cite{singh2020investigating}.
In Equation \eqref{eqn:min-max}, $\mathbf{X}$ represents the image to be normalized, $min$ and $max$ are the minimum and the maximum values computed on the overall dataset, $\varepsilon$ is set to $1\cdot10^{-6}$ and it is used to stabilize the division with a weak denominator, and lastly $\mathbf{Y}$ is the normalized image. The minimum and the maximum values have been computed on the overall dataset in order to restore the range of the model output using the inverse formula given by:

\begin{equation}
    \centering
    \mathbf{X} =  \mathbf{Y}[(max-min) + \varepsilon] + min
    \label{eqn:inv-min-max}
\end{equation}

The proposed model has been developed in \textit{Tensorflow-Keras} and trained, as summarized in Algorithm \ref{alg:training}, on the platform Google Colaboratory ~\cite{bisong2019google}, where each user can count on: $1)$ a GPU Tesla K80, having 2496 CUDA cores, compute 3.7, 12G GDDR5  VRAM, $2)$ a CPU single core hyper threaded i.e (1 core, 2 threads) Xeon Processors @2.3 Ghz (No Turbo Boost), $3)$ 45MB Cache, $4)$ 12.6 GB of available RAM and $5)$ 320 GB of available disk. 

The code has been made available open-source at the following link \cite{code} for further analysis and experiments.

\begin{algorithm}[!ht]
\SetAlgoLined
\textbf{initializeModel()};\\
\For{epoch$\leftarrow 0$ \KwTo epochs \KwBy $1$}{
    \Comment{Load Clean Image from dataset}
    groundTruth  = \textbf{loadFromTrainingSet}();\\
    \Comment{Add noise to Clean Image}
    inputImg =\\
    groundTruth * \textbf{generateNoiseGamma}(looks = 4);\\
    \Comment{Normalize data}
    groundTruth = \textbf{normalize}(groundTruth);\\
    inputImg = \textbf{normalize}(inputImg);\\
    \Comment{Apply CNN}
    residualImage = \textbf{applyCNN}(inputImg);\\
    rawFiltered = \textbf{subtract}(inputImg, residualImage);\\
    filtered = \textbf{sigmoid}(rawFiltered);\\
    \Comment{Update CNN weights}
    error, grad = \textbf{computeErrorGrad}(filtered, groundTruth);\\
    \textbf{updateWeights}(error, grad);\\
    }
\caption{Training of the proposed model}\label{alg:training}    
\end{algorithm}

\section{Results and Discussion}
This section will present the results obtained with the proposed model on non-SAR data, on the proposed dataset and on a real GRD image. It is important to specify that real GRD data are slightly different from simulated ones, because the speckle has a spatially variable correlation that is not considered by the Goodman's model. 

Moreover, testing the proposed model also on real GRD data demonstrated its good performances, even if trained only on simulated data (where log-scaling is not applied and so the statistical characterization of data is maintained).
 
Moreover, comparisons with other models have been carried out even if it is essential to underline that this process is always very complex and delicate. Indeed, the predictive ability of the deep neural networks depends not only on the model architecture, but also on the data used for training them, and it is worth to highlight that the works in the state of the art use different data sources (e.g. Sentinel-1 SLC, TerraSAR-X, etc.) and different statistical models for the speckle \cite{dalsasso2021sar2sar, molini2021speckle2void, cozzolino2020nonlocal}. Therefore, testing the algorithms on data, that are different from those used for training, is not an immediate process and could lead to inconsistencies in results. A further difference can be given by the type of data pre-processing or  the type of data being used (e.g. GRD, Single Look  (SLC), etc.), since also in these cases the comparison procedure could lead to imprecise results. Starting from these considerations, the model has been tested on the proposed dataset and also on a real GRD image,   downloaded with the tool presented in \cite{sebastianelli2020automatic} and made available on the GitHub repository \cite{code}.

Moreover,  additional experiments have been carried out by applying  the found state-of-the-art speckle filters to the created dataset and to the real GRD image,  by adopting the data pre-processing requested by each filter.

It seems honest to point out that, in some cases the available code is really hard to read and to use, and also the data pre-processing is made unclear. Therefore, the replicability of some results could be affected by these critical issues. Instead, in our case, among the objectives of our work, as already pointed out, we  have considered of high priority to set the feasibility of a code open access and user-friendly in order to make our work and the results replicable by all the interested researchers.

\subsection{Results on COCO dataset}

To prove the generalization of the proposed model, we decided to firstly train and validate it on a subset of the well known "Common objects in context (COCO)" dataset \cite{lin2014microsoft}.

We repeated the simulations by varying the type of noise and its power, and the results have been evaluated by using two metrics, the PSNR (Peak Signal-to-Noise Ratio) and the SSIM (Structural Similarity Index)  \cite{hore2010image}. 

The PSNR, in this case, offers a measure of the filtering process quality, and it is given by the ratio between the maximum power of the ground truth acquisition and the residual error between the ground truth and the acquisition filtered by the model, in accordance to the equation:
\begin{equation}
    \centering
    \begin{split}
    &\text{PSNR}(\mathbf{x}_{inputs}^v, \mathbf{x}_{ground\_ truths}^v)=\\&=10\log_{10}{\left( \frac{\max{(\mathbf{x}_{ground\_ truths}^v)}^2}{\|\mathbf{x}_{ground\_ truths}^v-\mathbf{x}_{inputs}^v\|_{2}^{2}/N}\right)}
    \end{split}
    \label{eqn:psnr}
\end{equation}

\begin{figure*}[!ht]
    \centering
    \resizebox{2\columnwidth}{!}{
    \begin{tabular}{ccccccccccc}
    Ground & Noisy & Model & Noisy & Model & Noisy & Model & Noisy  & Model & Noisy & Model\\
    
    Truth & Input & Prediction &  Input & Prediction & Input & Prediction & Input & Prediction & Input & Prediction\\
    
    & $\mu:0, \sigma:0.01$ &  &  $\mu:0, \sigma:0.02$ &  & $\mu:0, \sigma:0.05$ &  & $\mu:0, \sigma:0.08$ &  & $\mu:0, \sigma:0.1 $ & \\
    \midrule
    \includegraphics[width=0.18\columnwidth]{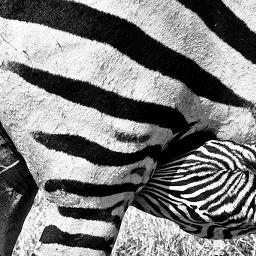}&
    \includegraphics[width=0.18\columnwidth]{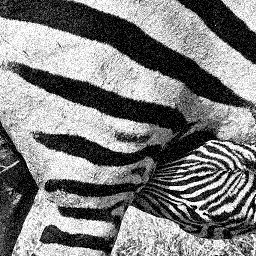}&
    \includegraphics[width=0.18\columnwidth]{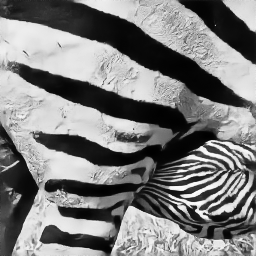}&
    \includegraphics[width=0.18\columnwidth]{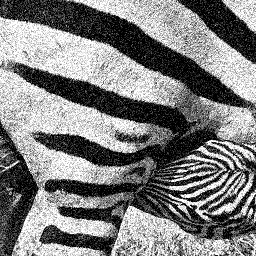}&
    \includegraphics[width=0.18\columnwidth]{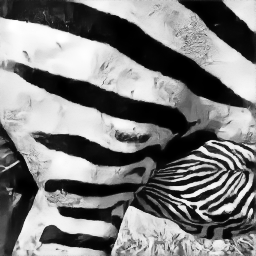}&
    \includegraphics[width=0.18\columnwidth]{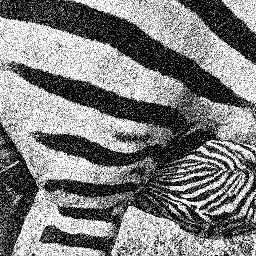}&
    \includegraphics[width=0.18\columnwidth]{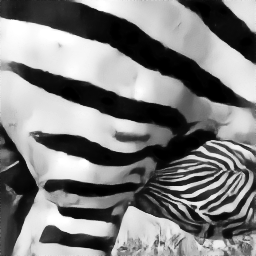}&
    \includegraphics[width=0.18\columnwidth]{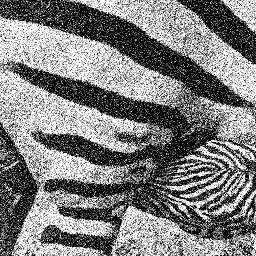}&
    \includegraphics[width=0.18\columnwidth]{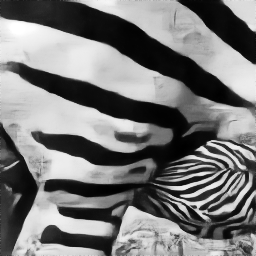}&
    \includegraphics[width=0.18\columnwidth]{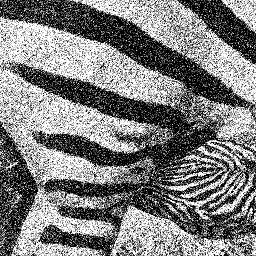}&
    \includegraphics[width=0.18\columnwidth]{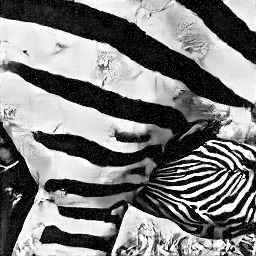}\\
    
    \includegraphics[width=0.18\columnwidth]{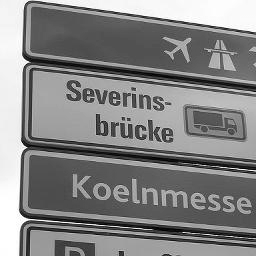}&
    \includegraphics[width=0.18\columnwidth]{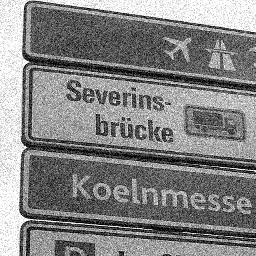}&
    \includegraphics[width=0.18\columnwidth]{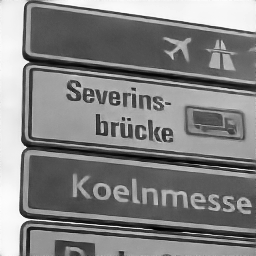}&
    \includegraphics[width=0.18\columnwidth]{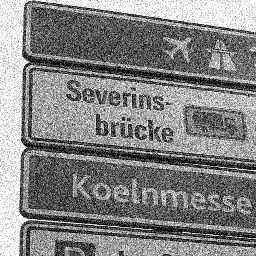}&
    \includegraphics[width=0.18\columnwidth]{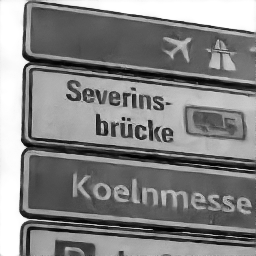}&
    \includegraphics[width=0.18\columnwidth]{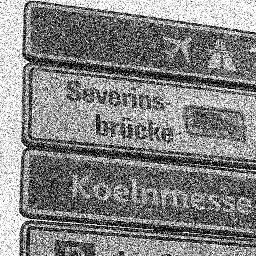}&
    \includegraphics[width=0.18\columnwidth]{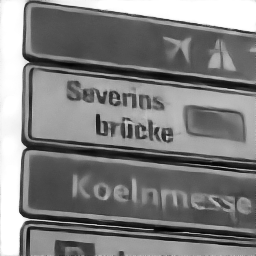}&
    \includegraphics[width=0.18\columnwidth]{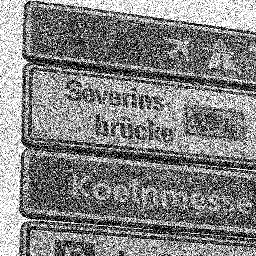}&
    \includegraphics[width=0.18\columnwidth]{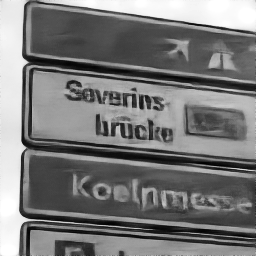}&
    \includegraphics[width=0.18\columnwidth]{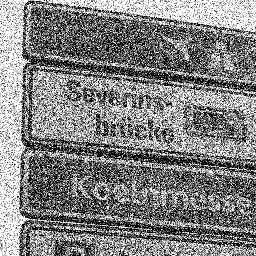}&
    \includegraphics[width=0.18\columnwidth]{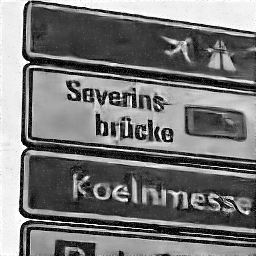}\\
    
    \includegraphics[width=0.18\columnwidth]{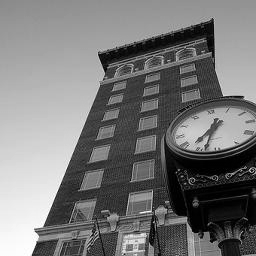}&
    \includegraphics[width=0.18\columnwidth]{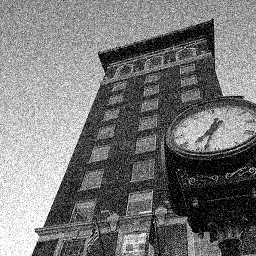}&
    \includegraphics[width=0.18\columnwidth]{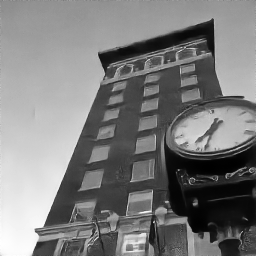}&
    \includegraphics[width=0.18\columnwidth]{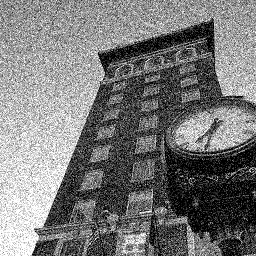}&
    \includegraphics[width=0.18\columnwidth]{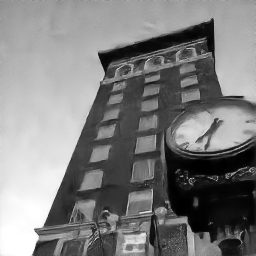}&
    \includegraphics[width=0.18\columnwidth]{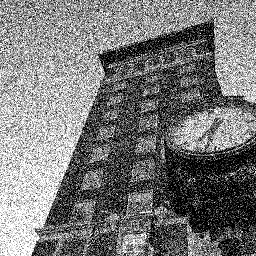}&
    \includegraphics[width=0.18\columnwidth]{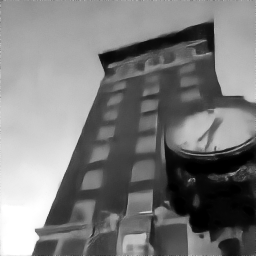}&
    \includegraphics[width=0.18\columnwidth]{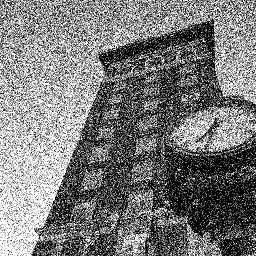}&
    \includegraphics[width=0.18\columnwidth]{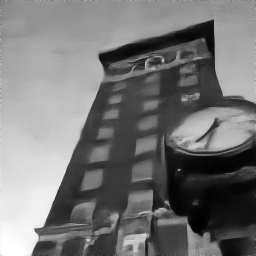}&
    \includegraphics[width=0.18\columnwidth]{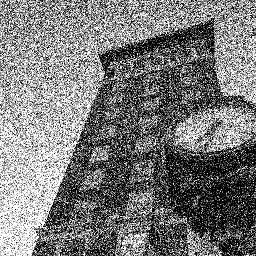}&
    \includegraphics[width=0.18\columnwidth]{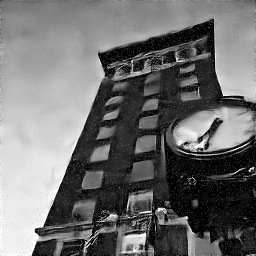}\\
    
    \includegraphics[width=0.18\columnwidth]{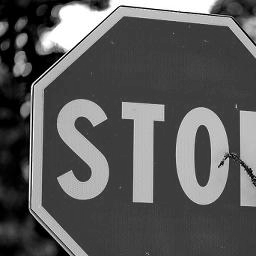}&
    \includegraphics[width=0.18\columnwidth]{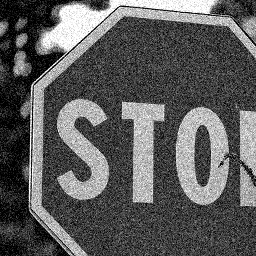}&
    \includegraphics[width=0.18\columnwidth]{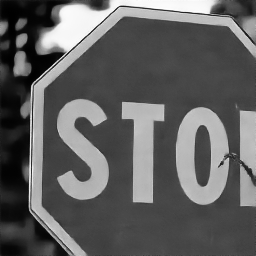}&
    \includegraphics[width=0.18\columnwidth]{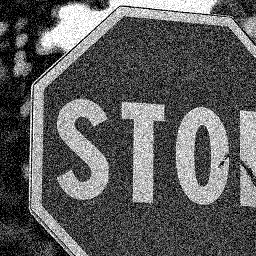}&
    \includegraphics[width=0.18\columnwidth]{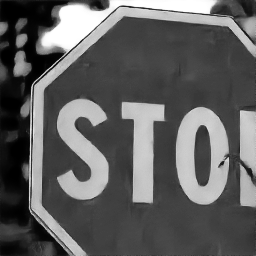}&
    \includegraphics[width=0.18\columnwidth]{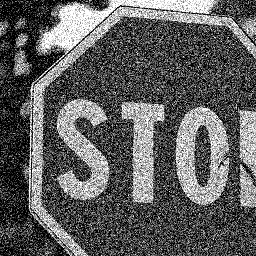}&
    \includegraphics[width=0.18\columnwidth]{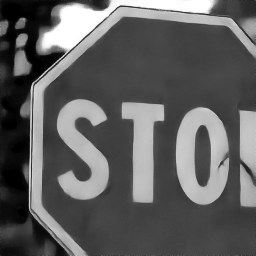}&
    \includegraphics[width=0.18\columnwidth]{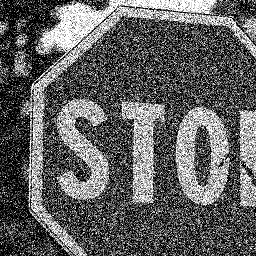}&
    \includegraphics[width=0.18\columnwidth]{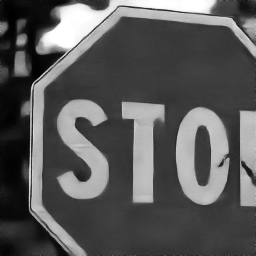}&
    \includegraphics[width=0.18\columnwidth]{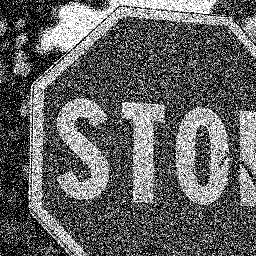}&
    \includegraphics[width=0.18\columnwidth]{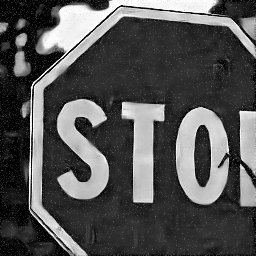}\\
    
    \includegraphics[width=0.18\columnwidth]{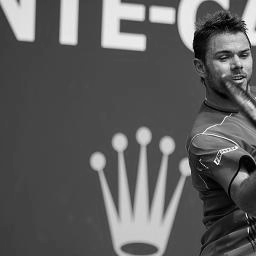}&
    \includegraphics[width=0.18\columnwidth]{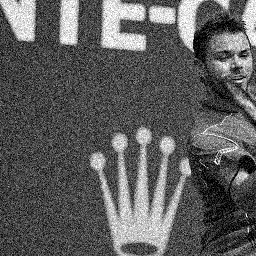}&
    \includegraphics[width=0.18\columnwidth]{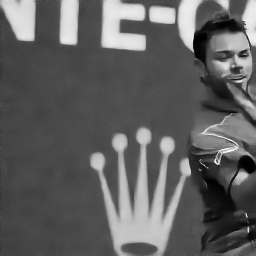}&
    \includegraphics[width=0.18\columnwidth]{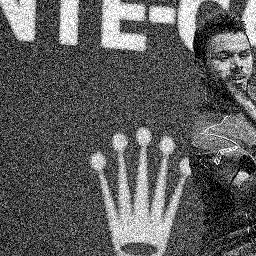}&
    \includegraphics[width=0.18\columnwidth]{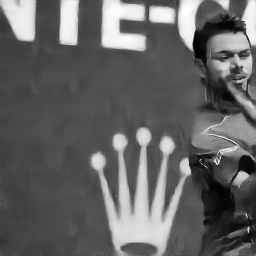}&
    \includegraphics[width=0.18\columnwidth]{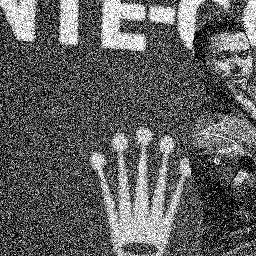}&
    \includegraphics[width=0.18\columnwidth]{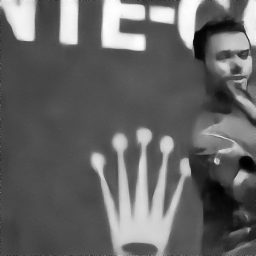}&
    \includegraphics[width=0.18\columnwidth]{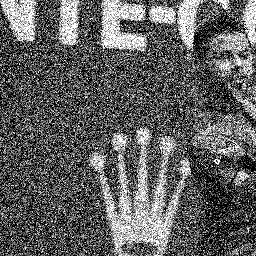}&
    \includegraphics[width=0.18\columnwidth]{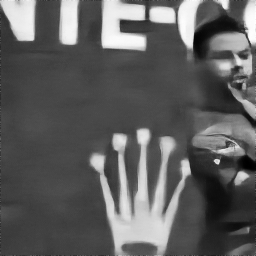}&
    \includegraphics[width=0.18\columnwidth]{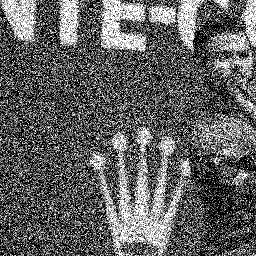}&
    \includegraphics[width=0.18\columnwidth]{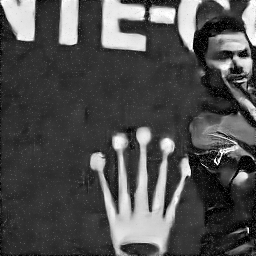}\\
    
    \includegraphics[width=0.18\columnwidth]{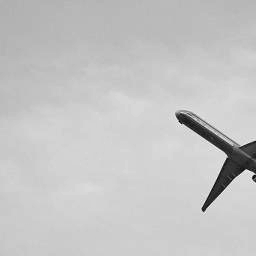}&
    \includegraphics[width=0.18\columnwidth]{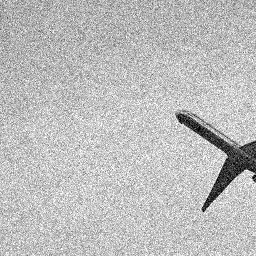}&
    \includegraphics[width=0.18\columnwidth]{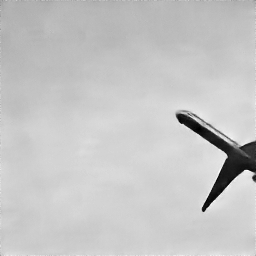}&
    \includegraphics[width=0.18\columnwidth]{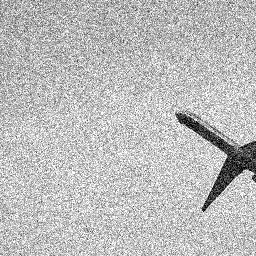}&
    \includegraphics[width=0.18\columnwidth]{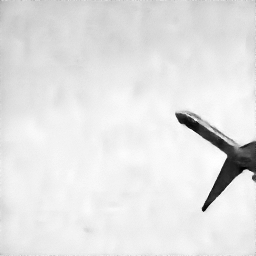}&
    \includegraphics[width=0.18\columnwidth]{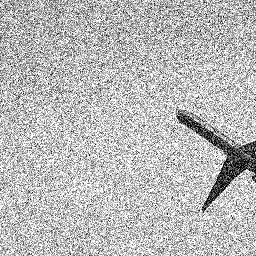}&
    \includegraphics[width=0.18\columnwidth]{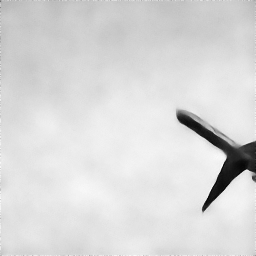}&
    \includegraphics[width=0.18\columnwidth]{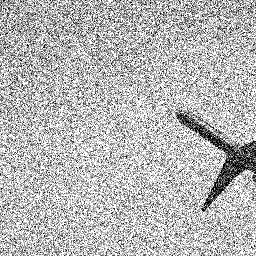}&
    \includegraphics[width=0.18\columnwidth]{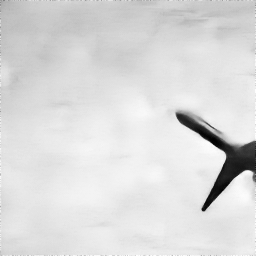}&
    \includegraphics[width=0.18\columnwidth]{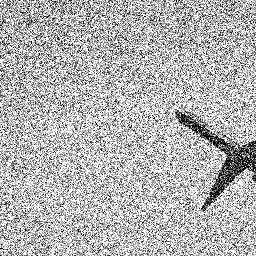}&
    \includegraphics[width=0.18\columnwidth]{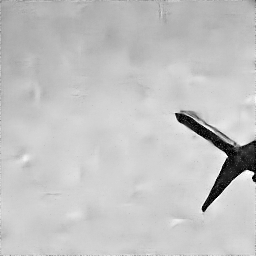}\\
    
    \includegraphics[width=0.18\columnwidth]{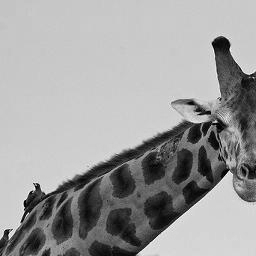}&
    \includegraphics[width=0.18\columnwidth]{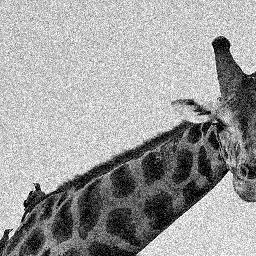}&
    \includegraphics[width=0.18\columnwidth]{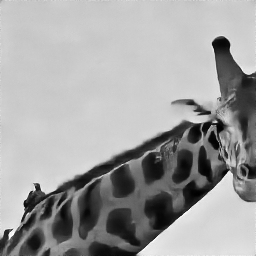}&
    \includegraphics[width=0.18\columnwidth]{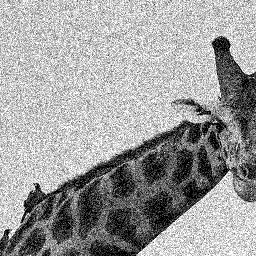}&
    \includegraphics[width=0.18\columnwidth]{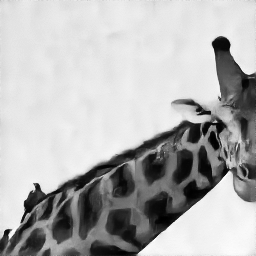}&
    \includegraphics[width=0.18\columnwidth]{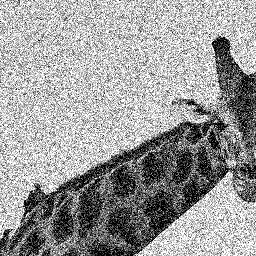}&
    \includegraphics[width=0.18\columnwidth]{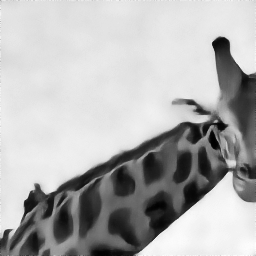}&
    \includegraphics[width=0.18\columnwidth]{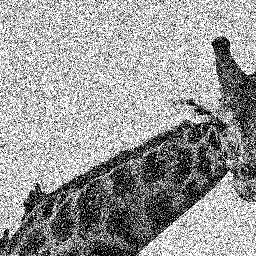}&
    \includegraphics[width=0.18\columnwidth]{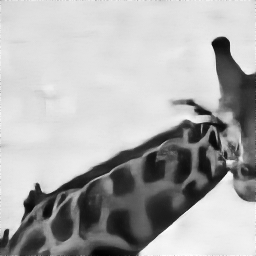}&
    \includegraphics[width=0.18\columnwidth]{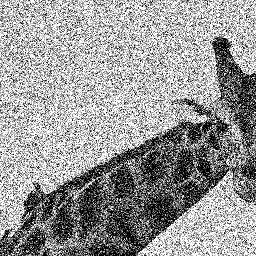}&
    \includegraphics[width=0.18\columnwidth]{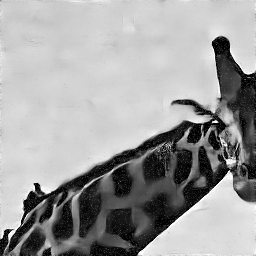}\\
    \end{tabular}
    }
    \caption{Graphical results of the proposed model tested by adding Gaussian noise to the gray-scale version of the COCO dataset. Several simulation were made by varying the standard deviation of the Gaussian noise while keeping the mean equal to zero.}
    \label{fig:nosarres1}
\end{figure*}

\vspace{0.3cm}
The \textit{SSIM} measures the similarity between two acquisitions and considers the image degradation as perceived change in the structural information. Let $\mathbf{x}$ be an input acquisition taken from the validation dataset $\Vec{\mathbf{X}}_{inputs}^v$ and let $\mathbf{y}$ be a ground truth acquisition taken from the validation dataset $\Vec{\mathbf{X}}_{ground\_ truths}^v$, then the \textit{SSIM} is computed by using the equation:
\begin{equation}
    \centering
    SSIM(x,y)=\frac{(2\mu_x\mu_y+c_1)(2\sigma_{xy}+c_2)}{(\mu_x^2+\mu_y^2+c_1)(\sigma_x^2+\sigma_y^2+c_2)}
    \label{eqn:ssim}
\end{equation}

where $\mu_x$ is the average of $\mathbf{x}$, $\mu_y$ is the average of $\mathbf{y}$, $\sigma_x^2$ is the variance of $\mathbf{x}$, $\sigma_y^2$ is the variance of $\mathbf{y}$, $\sigma_{xy}^2$ is the covariance of $\mathbf{x}$ and  $\mathbf{y}$, $c_1 = (k_1D)^2$ and $c_2 = (k_2D)^2$ are two variables that stabilize the division with a weak denominator, $D$ is the dynamic range of the pixel-values (typical $2^{\#BitsPerPixel}-1$), $k_1=0.01$ and $k_2=0.03$ by default.

\textbf{Gaussian Noise.} In a first phase,  experiments were run to test the proposed model when Gaussian noise had been added to the gray-scale version of the COCO dataset. Several simulation were made by varying the standard deviation of the Gaussian noise while keeping the mean equals to zero. For each simulation, training and testing of the model has been repeated. Graphical results are reported in Figure \ref{fig:nosarres1}, where each row represents a different test,  the first column represents the ground truth, and the other couples of columns represent the noisy input at a certain noise power and the model prediction, respectively. Quantitative results corresponding to Figure \ref{fig:nosarres1}, are reported in Table \ref{tab:nosarres1}. The first columns indicates the standard deviation of the Gaussian noise, the second one the metric used to evaluate the results and the remaining ones report the metric values for each test. Note that the PSNR and SSIM are computed both between the ground truth (GT) and noisy input (N) and the GT and the model prediction (P). Results indicate that the proposed model is able to produce good qualitative and quantitative results, indeed the PSNR and SSIM measured between  GT and P are always higher than the ones measured between GT and N.

\begin{table}[!ht]
    \centering
    \resizebox{\columnwidth}{!}{\begin{tabular}{llccccccc}
    \toprule
    Std. &   Metric      & Test 1 & Test 2 & Test 3 & Test 4 & Test 5 & Test 6 & Test 7\\
    \midrule
          & PSNR(GT, N)  & 21.12 & 20.35 & 20.35 & 20.38 & 20.12 & 20.10 & 20.22\\
          & PSNR(GT, P)  & 22.75 & 27.78 & 26.28 & 28.57 & 30.11 & 25.53 & 29.28\\
    0.01  & SSIM(GT, N)  &  0.74 &  0.62 &  0.51 &  0.48 &  0.45 &  0.30 &  0.44\\
          & SSIM(GT, P)  &  0.82 &  0.96 &  0.90 &  0.95 &  0.96 &  0.98 &  0.95\\
    \midrule
          & PSNR(GT, N) & 18.36 & 17.46 & 17.56 & 17.59 & 17.36 & 17.35 & 17.56 \\
          & PSNR(GT, P) & 21.82 & 25.59 & 24.41 & 27.85 & 26.38 & 17.01 & 20.42 \\
    0.02  & SSIM(GT, N) &  0.63 &  0.51 &  0.39 &  0.37 &  0.33 &  0.20 &  0.33 \\
          & SSIM(GT, P) &  0.77 &  0.93 &  0.87 &  0.94 &  0.94 &  0.96 &  0.92 \\
    \midrule
          & PSNR(GT, N) & 14.81 & 13.89 & 14.02 & 14.25 & 13.91 & 14.08 & 14.24 \\
          & PSNR(GT, P) & 20.01 & 23.42 & 23.86 & 26.89 & 25.25 & 18.06 & 20.19 \\
    0.05  & SSIM(GT, N) &  0.48 &  0.38 &  0.24 &  0.25 &  0.21 &  0.11 &  0.21 \\
          & SSIM(GT, P) &  0.68 &  0.89 &  0.82 &  0.93 &  0.92 &  0.97 &  0.91 \\
    \midrule
          & PSNR(GT, N) & 13.04 & 12.28 & 12.39 & 12.65 & 12.41 & 12.53 & 12.64 \\
          & PSNR(GT, P) & 19.84 & 22.60 & 23.48 & 26.36 & 25.06 & 19.89 & 20.56 \\
    0.08  & SSIM(GT, N) &  0.40 &  0.30 &  0.20 &  0.21 &  0.17 &  0.10 &  0.16 \\
          & SSIM(GT, P) &  0.68 &  0.88 &  0.80 &  0.92 &  0.91 &  0.96 &  0.89 \\
    \midrule
          & PSNR(GT, N) & 12.23 & 11.54 & 11.69 & 11.84 & 11.60 & 11.76 & 11.88 \\
          & PSNR(GT, P) & 19.16 & 20.20 & 21.72 & 23.12 & 22.31 & 24.01 & 22.13 \\
    0.1   & SSIM(GT, N) &  0.36 &  0.27 &  0.16 &  0.18 &  0.14 &  0.07 &  0.14 \\
          & SSIM(GT, P) &  0.66 &  0.81 &  0.74 &  0.87 &  0.84 &  0.79 &  0.77 \\
    \bottomrule
    \end{tabular}}
    \caption{Quantitative results for experiments reported in Figure \ref{fig:nosarres1}. The first columns indicates the standard deviation of the Gaussian noise, the second one the metric used to evaluate the results and the remaining ones report the metric values for each test. Please note that the PSRN and SSIM are computed both between the ground truth (GT) and noisy input (N) and the GT and the model prediction (P).}
    \label{tab:nosarres1}
\end{table}

\begin{figure*}[!ht]
    \centering
    \resizebox{1.7\columnwidth}{!}{
    \begin{tabular}{ccccccc}
    Ground & Noisy & Model & Noisy & Model & Noisy & Model\\
    
    Truth & Input & Prediction &  Input & Prediction & Input & Prediction \\
    
    & $L:1$  &  & $L:2$ &  & $L:4$ &  \\
    
    \midrule
    \includegraphics[width=0.18\columnwidth]{figures/coco/b_0_gt-l_1.png}& 
    \includegraphics[width=0.18\columnwidth]{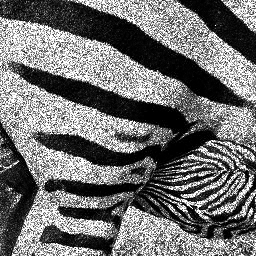}&
    \includegraphics[width=0.18\columnwidth]{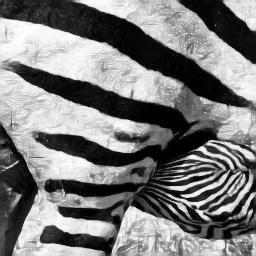}&
    \includegraphics[width=0.18\columnwidth]{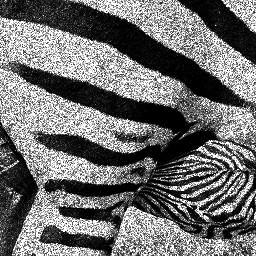}&
    \includegraphics[width=0.18\columnwidth]{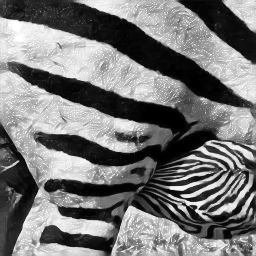}&
    \includegraphics[width=0.18\columnwidth]{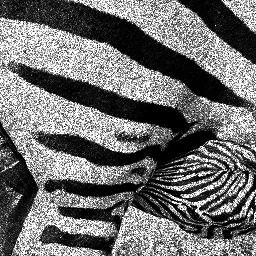}&
    \includegraphics[width=0.18\columnwidth]{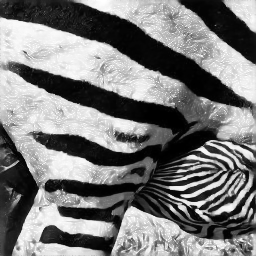}\\
    
    \includegraphics[width=0.18\columnwidth]{figures/coco/b_1_gt-l_1.png}& 
    \includegraphics[width=0.18\columnwidth]{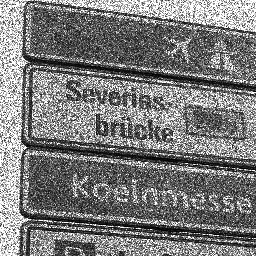}&
    \includegraphics[width=0.18\columnwidth]{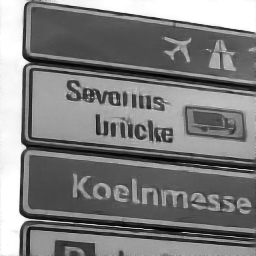}&
    \includegraphics[width=0.18\columnwidth]{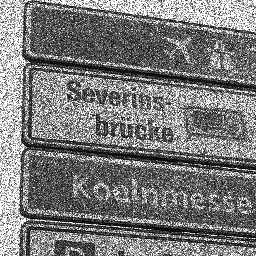}&
    \includegraphics[width=0.18\columnwidth]{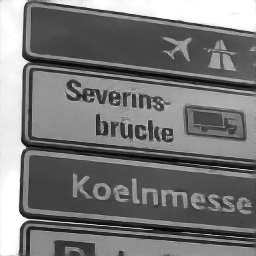}&
    \includegraphics[width=0.18\columnwidth]{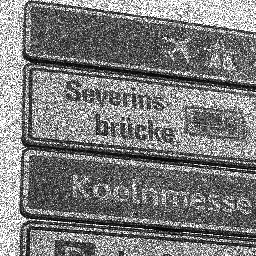}&
    \includegraphics[width=0.18\columnwidth]{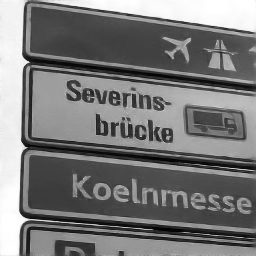}\\
    
    \includegraphics[width=0.18\columnwidth]{figures/coco/b_2_gt-l_1.png}& 
    \includegraphics[width=0.18\columnwidth]{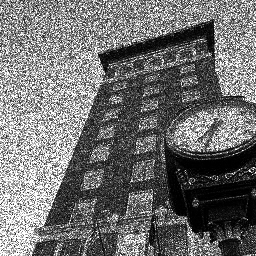}&
    \includegraphics[width=0.18\columnwidth]{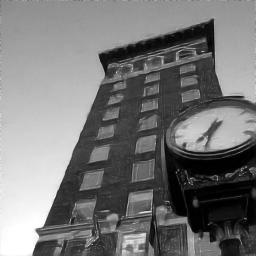}&
    \includegraphics[width=0.18\columnwidth]{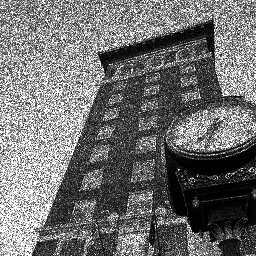}&
    \includegraphics[width=0.18\columnwidth]{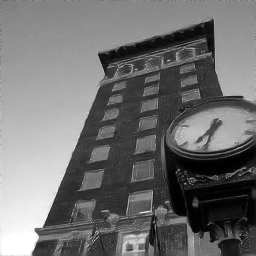}&
    \includegraphics[width=0.18\columnwidth]{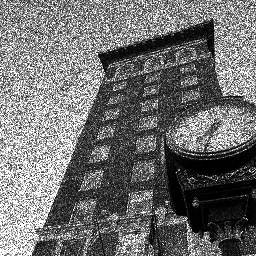}&
    \includegraphics[width=0.18\columnwidth]{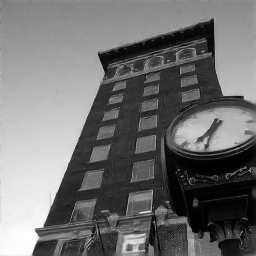}\\

    \includegraphics[width=0.18\columnwidth]{figures/coco/b_3_gt-l_1.png}& 
    \includegraphics[width=0.18\columnwidth]{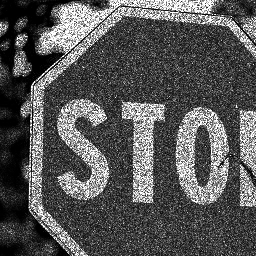}&
    \includegraphics[width=0.18\columnwidth]{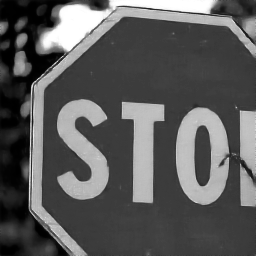}&
    \includegraphics[width=0.18\columnwidth]{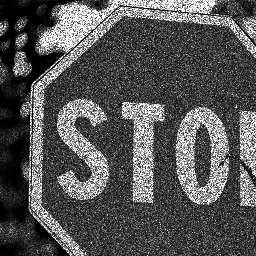}&
    \includegraphics[width=0.18\columnwidth]{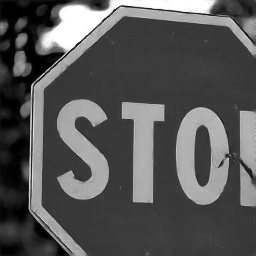}&
    \includegraphics[width=0.18\columnwidth]{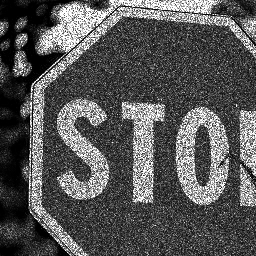}&
    \includegraphics[width=0.18\columnwidth]{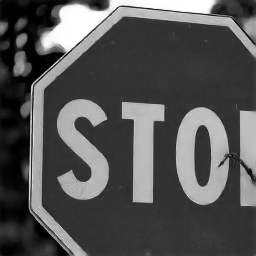}\\
    
    \includegraphics[width=0.18\columnwidth]{figures/coco/b_4_gt-l_1.png}& 
    \includegraphics[width=0.18\columnwidth]{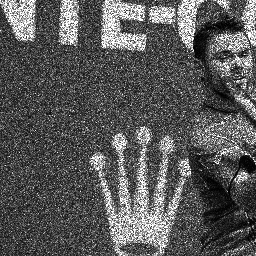}&
    \includegraphics[width=0.18\columnwidth]{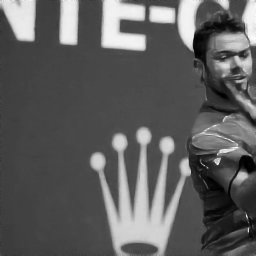}&
    \includegraphics[width=0.18\columnwidth]{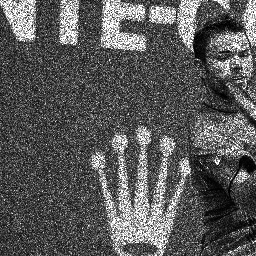}&
    \includegraphics[width=0.18\columnwidth]{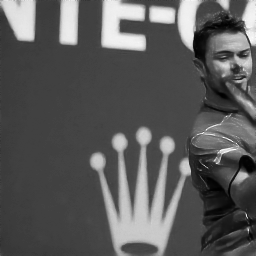}&
    \includegraphics[width=0.18\columnwidth]{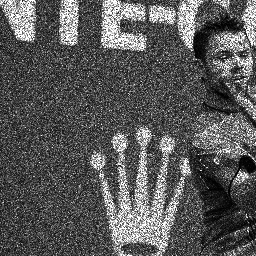}&
    \includegraphics[width=0.18\columnwidth]{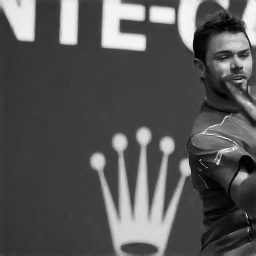}\\
    
    \includegraphics[width=0.18\columnwidth]{figures/coco/b_5_gt-l_1.png}& 
    \includegraphics[width=0.18\columnwidth]{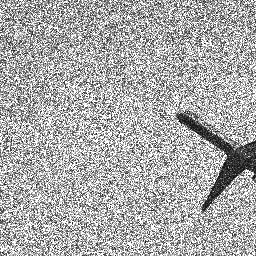}&
    \includegraphics[width=0.18\columnwidth]{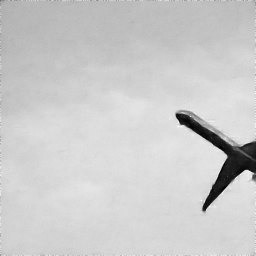}&
    \includegraphics[width=0.18\columnwidth]{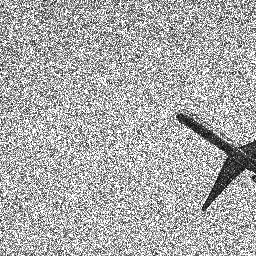}&
    \includegraphics[width=0.18\columnwidth]{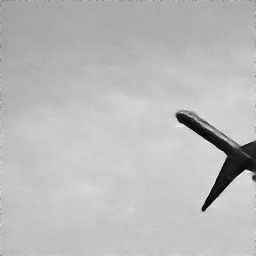}&
    \includegraphics[width=0.18\columnwidth]{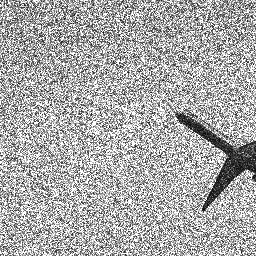}&
    \includegraphics[width=0.18\columnwidth]{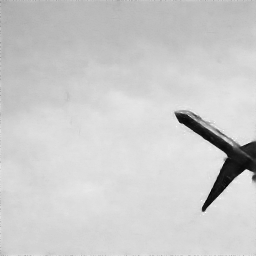}\\

    \includegraphics[width=0.18\columnwidth]{figures/coco/b_6_gt-l_1.png}& 
    \includegraphics[width=0.18\columnwidth]{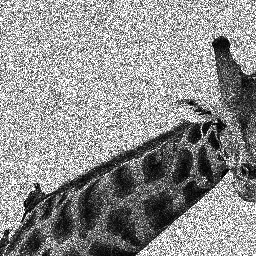}&
    \includegraphics[width=0.18\columnwidth]{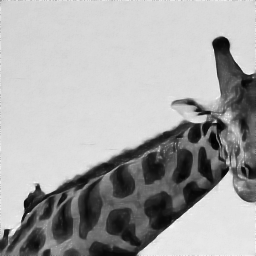}&
    \includegraphics[width=0.18\columnwidth]{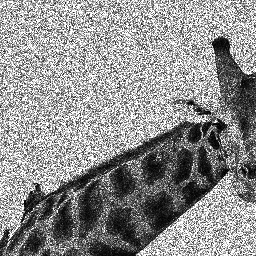}&
    \includegraphics[width=0.18\columnwidth]{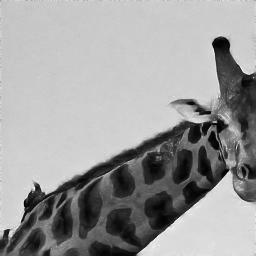}&
    \includegraphics[width=0.18\columnwidth]{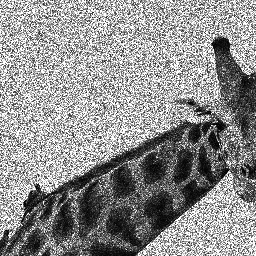}&
    \includegraphics[width=0.18\columnwidth]{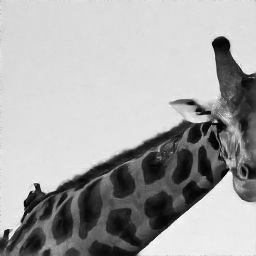}\\
    \end{tabular}
    }
    \caption{Graphical results of the proposed model tested by adding the simulated Speckle noise to the gray-scale version of the COCO dataset. Several simulation were made by varying the number of Looks.}
    \label{fig:nosarres2}
\end{figure*}

\textbf{Speckle Noise:} In a second phase, experiments were run to test the proposed  model by multiplying the speckle noise, as generated with the statistical model based on the Gamma distribution introduced before,  to the gray-scale version of the COCO dataset. Several simulations were made by varying the number of Looks. For each simulation the training and testing of the model has been repeated. Graphical results are reported in Figure \ref{fig:nosarres2}, where each rows represent a different test and each couple of columns, apart from the first column   showing  the ground truth, represent the noisy input at a certain look L and the model prediction, respectively. Quantitative results corresponding to Figure \ref{fig:nosarres2}, are reported in Table \ref{tab:nosarres2}. The first columns indicates the number of look of the speckle, the second one the metric used to evaluate the results and the remaining ones report the metric values for each test. Please note that PSNR and SSIM are computed both between the ground truth (GT) and the noisy input (N) and the GT and the model prediction (P). Results indicate also in this case that the proposed model is able to produce good qualitative and quantitative results, indeed the PSNR and SSIM measured between GT and P are always higher than the ones measured between GT and N.

\begin{figure*}[!ht]
    \centering
    \resizebox{2\columnwidth}{!}{
    \begin{tabular}{cc}
    \textbf{(\small Test 1)} & \includegraphics[width=2\columnwidth]{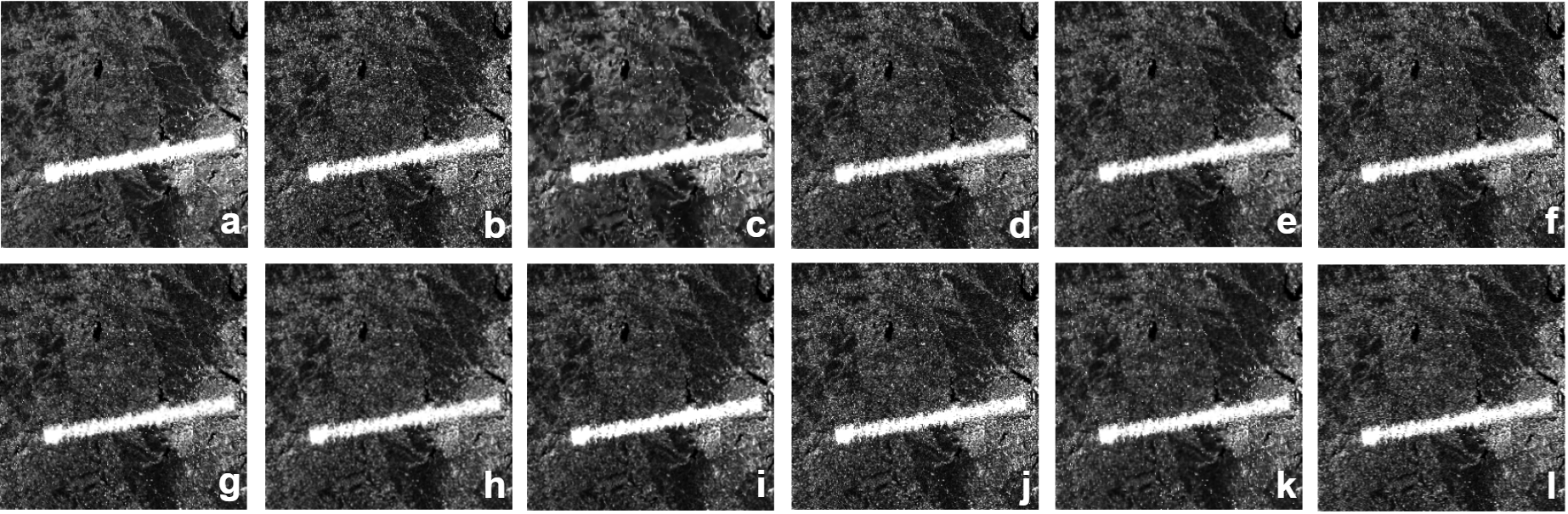}\\[3pt]
    \end{tabular}}
    \resizebox{2\columnwidth}{!}{
    \begin{tabular}{cc}
    \textbf{(\small Test 2)} & \includegraphics[width=2\columnwidth]{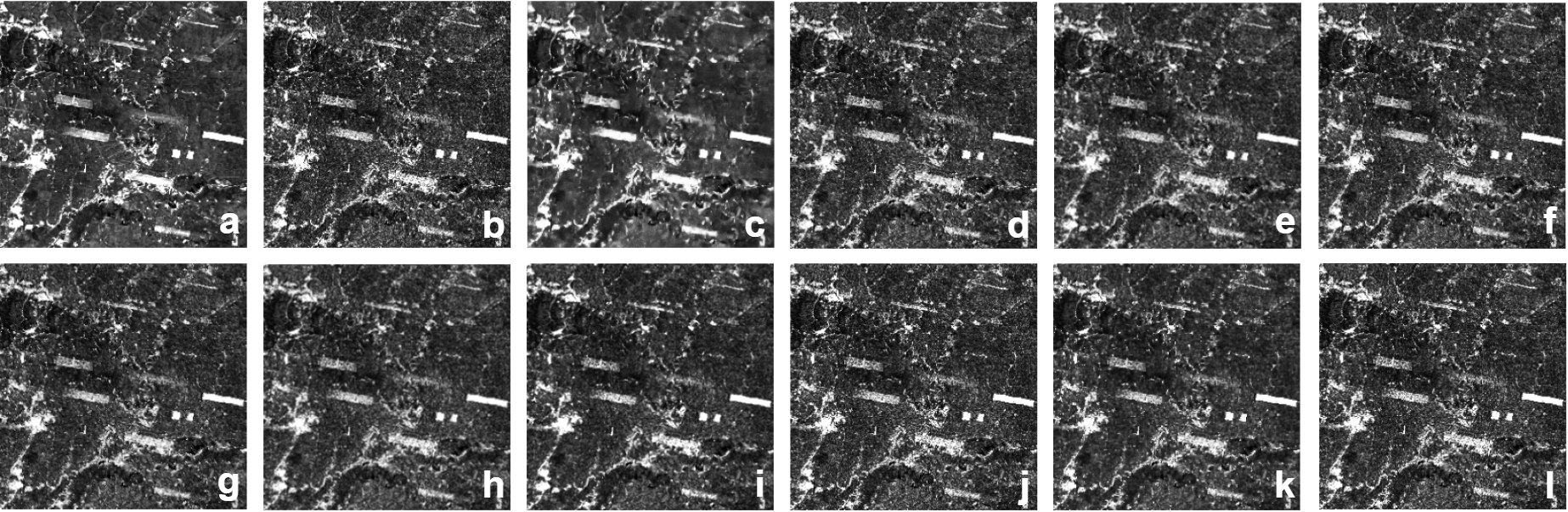}\\[3pt]
    \end{tabular}}
    \resizebox{2\columnwidth}{!}{
    \begin{tabular}{cc}
    \textbf{(\small Test 3)} & \includegraphics[width=2\columnwidth]{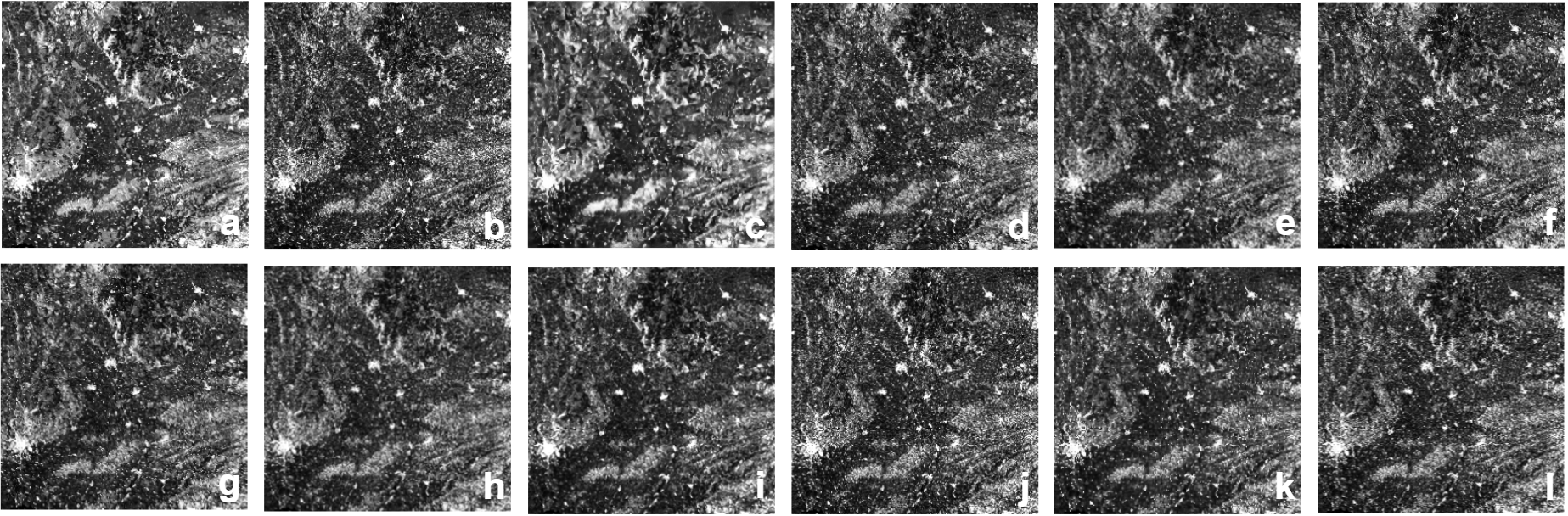}\\[3pt]
    \end{tabular}}
    \caption{Qualitative results on the testing dataset: (a) Ground Truth, (b) Input with speckle, (c) Proposed Model Prediction, (d) Lee , (e) Lee Enhanced, (f) Kuan, (g) Frost, (h) Mean, (i) Median, (j) Fastnl, (k) Bilateral, (l) SAR-BM3D}
    \label{fig:qualresults}
\end{figure*}

\begin{table}[!ht]
    \centering
    \resizebox{\columnwidth}{!}{\begin{tabular}{llccccccc}
    \toprule
    Looks &   Metric    & Test 1 & Test 2 & Test 3 & Test 4 & Test 5 & Test 6 & Test 7\\
    \midrule
          & PSNR(GT, N)  & 10.06 & 10.57 & 11.13 & 12.82 & 13.21 &  8.27 &  9.20\\
          & PSNR(GT, P)  & 20.59 & 25.17 & 26.14 & 28.19 & 30.44 & 27.29 & 23.19\\
    1     & SSIM(GT, N)  &  0.48 &  0.32 &  0.33 &  0.37 &  0.31 &  0.06 &  0.21\\
          & SSIM(GT, P)  &  0.77 &  0.94 &  0.90 &  0.96 &  0.97 &  0.98 &  0.95\\
    \midrule
          & PSNR(GT, N) & 12.14 & 12.71 & 13.29 & 14.99 & 15.38 & 10.42 & 11.35 \\
          & PSNR(GT, P) & 20.71 & 23.34 & 25.18 & 27.77 & 29.23 & 31.94 & 28.80 \\
    2     & SSIM(GT, N) &  0.58 &  0.43 &  0.41 &  0.48 &  0.41 &  0.09 &  0.27 \\
          & SSIM(GT, P) &  0.80 &  0.95 &  0.92 &  0.97 &  0.97 &  0.98 &  0.96 \\
    \midrule
          & PSNR(GT, N) & 14.65 & 15.17 & 15.72 & 17.36 & 17.75 & 12.82 & 13.82 \\
          & PSNR(GT, P) & 21.85 & 20.07 & 25.67 & 27.19 & 26.74 & 25.00 & 22.74 \\
    4     & SSIM(GT, N) &  0.68 &  0.53 &  0.50 &  0.58 &  0.53 &  0.14 &  0.34 \\
          & SSIM(GT, P) &  0.82 &  0.95 &  0.92 &  0.96 &  0.97 &  0.98 &  0.96 \\
    \bottomrule
    \end{tabular}}
    \caption{Quantitative results for experiments reported in Figure \ref{fig:nosarres2}. The first columns indicates the number of look of the speckle, the second one the metric used to evaluate the results and the remaining ones report the metric values for each test. Please note that the PSRN and SSIM are computed both between the ground truth (GT) and noisy input (N) and the GT and the model prediction (P).}
    \label{tab:nosarres2}
\end{table}

\subsection{Results and comparisons on the reference dataset}

The proposed model has been then evaluated through the \textit{PSNR} and  \textit{SSIM} metrics on the testing dataset  $\Vec{\mathbf{X}}_{inputs}^v, \Vec{\mathbf{X}}_{ground\_ truths}^v$, whose creation has been described in the previous subsection A. Dataset creation for the proposed CNN. The procedure  is summarized in Algorithm \ref{alg:testing}.

\begin{algorithm}[!ht]
\SetAlgoLined
model = \textbf{loadTrainedModel()}\\
\For{img$\leftarrow$ \KwTo testing dataset}{
    \Comment{Load Clean Image from dataset}
    groundTruth  = \textbf{loadFromTestingSet}();\\
    \Comment{Add noise to Clean Image}
    inputImg =\\
    ~~~groundTruth * \textbf{generateNoiseGamma}(looks=4);\\
    \Comment{Normalize data}
    inputImg = \textbf{normalize}(inputImg);\\
    \Comment{Apply Trained Model}
    prediction = \textbf{applyTrainedModel}(model, inputImg)\\
    prediction = \textbf{normalize.inverse}(\\
    ~~~~~~~~~~~\textbf{applyTrainedModel}(model, inputImg))\\
    \Comment{Save results}
    predictions.append(prediction);\\
    groundTruths.append(groundTruth);
    }
\Comment{Get scores}
psnr, ssim = \textbf{calculateScores}(groundTruths, predictions);

\caption{Evaluation of the proposed model on the testing dataset}\label{alg:testing}    
\end{algorithm}

Quantitative results have been reported in Table \ref{tab:results1}, where the averaged scores,   computed on the testing dataset, are shown to compare the Proposed Model Prediction (c)   with:  (a) the Ground Truth, (b) Input with speckle,  and (d-l) some  state-of-the-art speckle filters: Lee, Lee Enhanced, Kuan, Frost, Mean, Median, Fastnl, Bilateral, SAR-BM3D, \cite{ref_forTableIV,liu2013nonlocal, tomasi1998bilateral, balocco2010srbf,parrilli2011nonlocal}. The implementation of the speckle filters used above can be found at the following link \cite{erdogant2020findpeaks}.

Note that the \textit{SSIM} ranges from $0$ to $1$, where $1$ is the best/ideal value and for the \textit{PSNR}, being a signal to noise ratio, the greater the value the better the result, and therefore the ideal value is $+\infty$. It is possible to see from the table that our model outperforms all the others. Qualitative results have been reported in Figure \ref{fig:qualresults}, where three different test images have been used to verify the model. Corresponding ENL (Equivalent Number of Looks) values are reported in Table \ref{tab:enelquaresults}, with the ENL defined as:

\begin{equation}
    ENL =  \frac{\mu^2_x}{\sigma^2_x}
\end{equation}

where $\mu_x$ is the mean intensity for a homogeneous region and $\sigma_x$ is the standard deviation.

\begin{table}[!ht]
    \centering
    \begin{tabular}{lccc}
    \toprule
    \textbf{Model} & \multicolumn{3}{c}{\textbf{ENL}}\\
                   & \textbf{Test 1} & \textbf{Test 2} & \textbf{Test 3}\\
    \midrule
    \textbf{Proposed}  & \textbf{36.04} & \textbf{47.22} & \textbf{32.25} \\
    \textbf{Lee}                        & 17.73 & 10.70 & 10.52 \\
    \textbf{Lee Enhanced}               & 20.97 & 16.15 & 14.37 \\
    \textbf{Kuan}                       & 18.10 & 11.12 & 10.84 \\
    \textbf{Frost}                      & 19.49 & 12.81 & 12.24 \\
    \textbf{Mean}                       & 21.46 & 17.07 & 14.54 \\
    \textbf{Median}                     & 19.26 & 15.48 & 11.96 \\
    \textbf{Fastnl}                     & 12.64 & 17.95 &  7.59 \\
    \textbf{Bilateral}                  & 18.30 & 13.40 & 10.37 \\
    \textbf{SAR-BM3D}                  & 14.64 & 17.95 & 17.59 \\
    \bottomrule
    \end{tabular}
    \caption{ENL for tests in Figure \ref{fig:qualresults}.}
    \label{tab:enelquaresults}
\end{table}

\begin{table}[!ht]
    \centering
    \begin{tabular}{lcc}
    \toprule
    \textbf{Model} & PSNR & SSIM\\
    \midrule
    \textbf{Ground Truth (a)}     & $+\infty$ & 1.0\\
    \textbf{Speckled (b)}         & 15.70 & 0.58 \\
    \textbf{Proposed (c)}         & \textbf{19.21} & \textbf{0.75} \\
    \textbf{Lee (d)}              & 16.64 & 0.61 \\
    \textbf{Lee Enhanced (e)}     & 16.46 & 0.55 \\
    \textbf{Kuan (f)}             & 16.78 & 0.61 \\
    \textbf{Frost (g)}            & 16.93 & 0.61 \\
    \textbf{Mean (h)}             & 15.93 & 0.50 \\
    \textbf{Median (i)}           & 15.39 & 0.48 \\
    \textbf{Fastnl (j)}           & 15.67 & 0.58 \\
    \textbf{Bilateral (k)}        & 17.29 & 0.62 \\
    \textbf{SAR-BM3D (l)}         & 15.67 & 0.58 \\
    \bottomrule
    \end{tabular}
    \caption{
    Average scores for the proposed model computed on the testing dataset: (a) Ground Truth, (b) Input with speckle, (c) Proposed Model Prediction, (d) Lee, (e) Lee Enhanced, (f) Kuan, (g) Frost, (h) Mean , (i) Median , (j) Fastnl, (k) Bilateral, (l) SAR-BM3D}
    \label{tab:results1}
\end{table}

\subsection{Results and comparisons on a real GRD image}
The real GRD image has been downloaded from the Copernicus Open Access Hub \cite{acceshub}. The motivation behind using also this particular image for testing the model depends mainly on two factors: $1)$ to give a further example of how the model performs on different source of data; and $2)$ to give interested readers an alternative procedure to use the model.

The same pre-processing of GEE data has been applied through the SNAP software \cite{snap}. This pre-processing is needed to bring the image under testing to the same conditions of the dataset. The steps involved in this phase are:

\begin{enumerate}
    \item \textbf{Apply orbit file}: updates orbit metadata with a restituted orbit file (or a precise orbit file if the restituted one is not available).
    \item \textbf{GRD border noise removal}: removes low intensity noise and invalid data on scene edges.
    \item \textbf{Thermal noise removal}: removes additive noise in sub-swaths to help reduce discontinuities between sub-swaths for scenes in multi-swath acquisition modes.
    \item \textbf{Radiometric calibration}: computes backscattered intensity using sensor calibration parameters in the GRD metadata.
    \item\textbf{ Terrain correction} (orthorectification): converts data from ground range geometry, which does not take terrain into account, to $\sigma^\circ$ using the SRTM 30 meter DEM or the ASTER DEM for high latitudes (greater than $60^\circ$ or less than $-60^\circ$).
    \item \textbf{Subset}: selects a portion of the input image.
\end{enumerate}

In this scenario, where the ground truth is not available, the only way to measure the  model performance is through the ENL. The testing of the proposed model on the real GRD image is summarized in Algorithm \ref{alg:testing2}.

\begin{algorithm}[!ht]
\SetAlgoLined
model = \textbf{loadTrainedModel()}\\
\Comment{Load real GRD Image}
inputImg  = \textbf{loadImage}();\\
\Comment{Normalize data}
inputImg = \textbf{normalize}(inputImg);\\
\Comment{Apply Trained Model}
prediction = \textbf{normalize.inverse}(\\
~~~~~~~~~~~\textbf{applyTrainedModel}(model, inputImg))\\
\Comment{Get scores}
enl = \textbf{calculateScore}(prediction);
\caption{Testing of the proposed model on the real GRD image}\label{alg:testing2}    
\end{algorithm}

\begin{figure*}[!ht]
    \centering
    \resizebox{2\columnwidth}{!}{
    \begin{tabular}{ccc}
    \includegraphics[width=0.60\columnwidth]{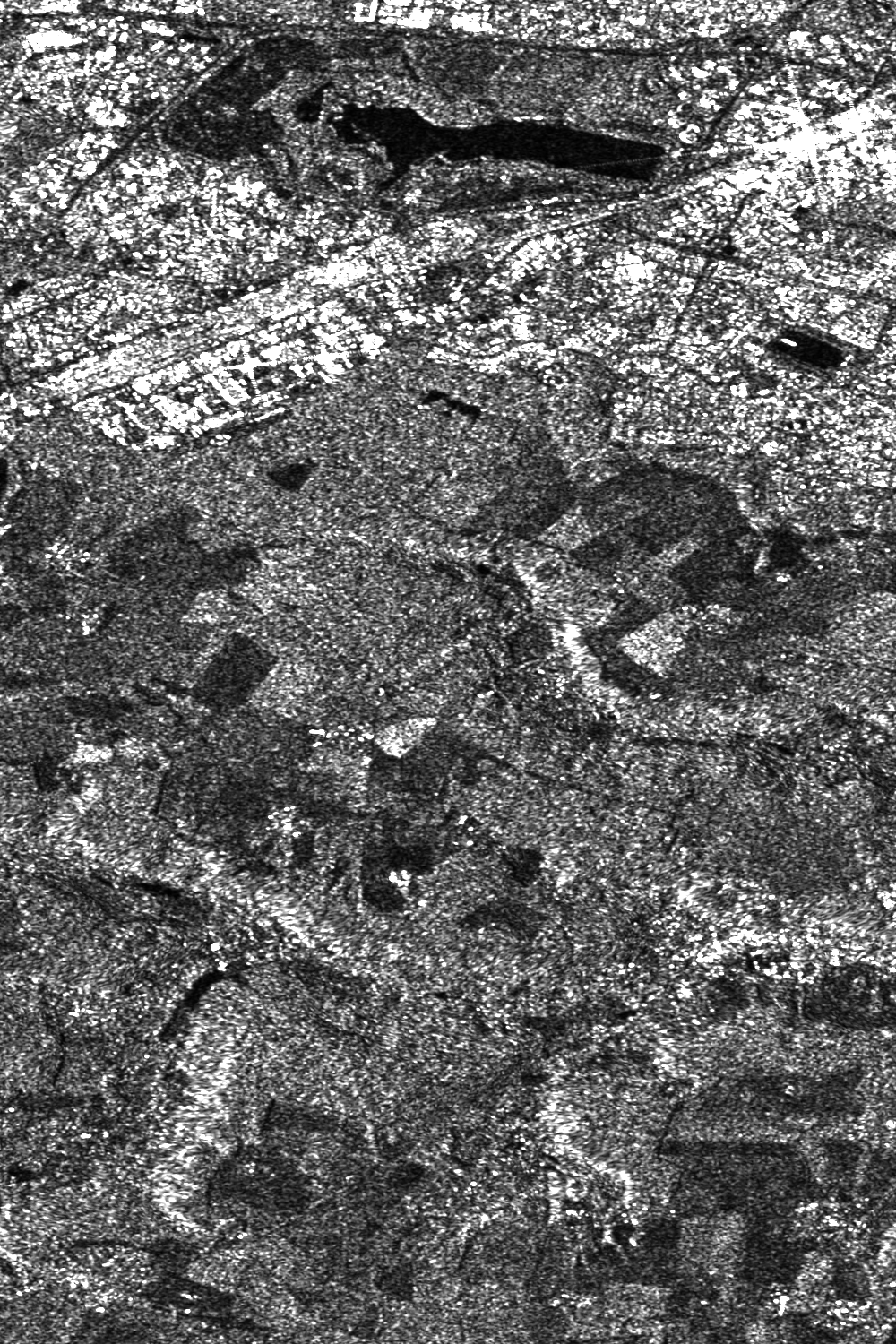} &
    \includegraphics[width=0.60\columnwidth]{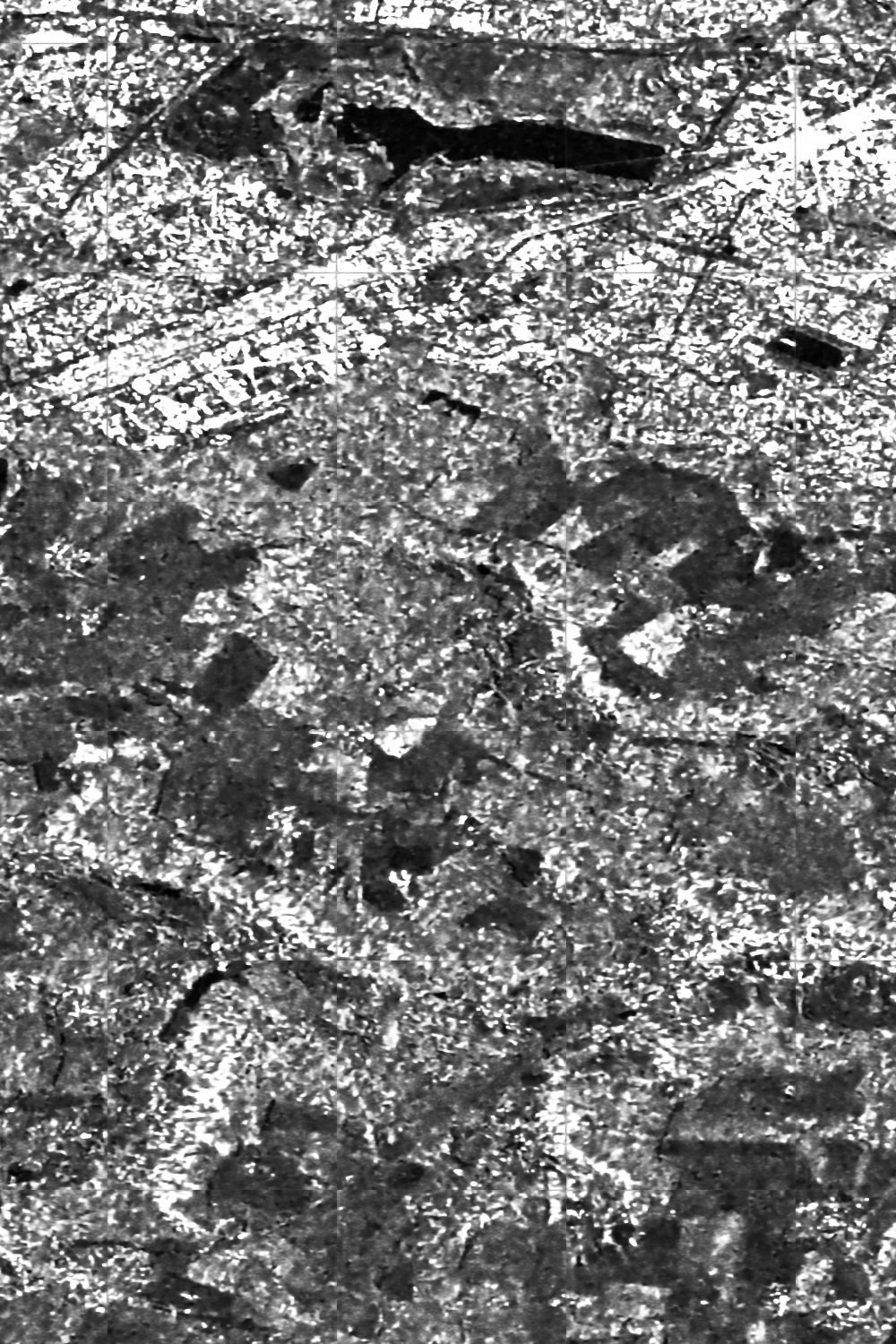} & 
    \includegraphics[width=0.60\columnwidth]{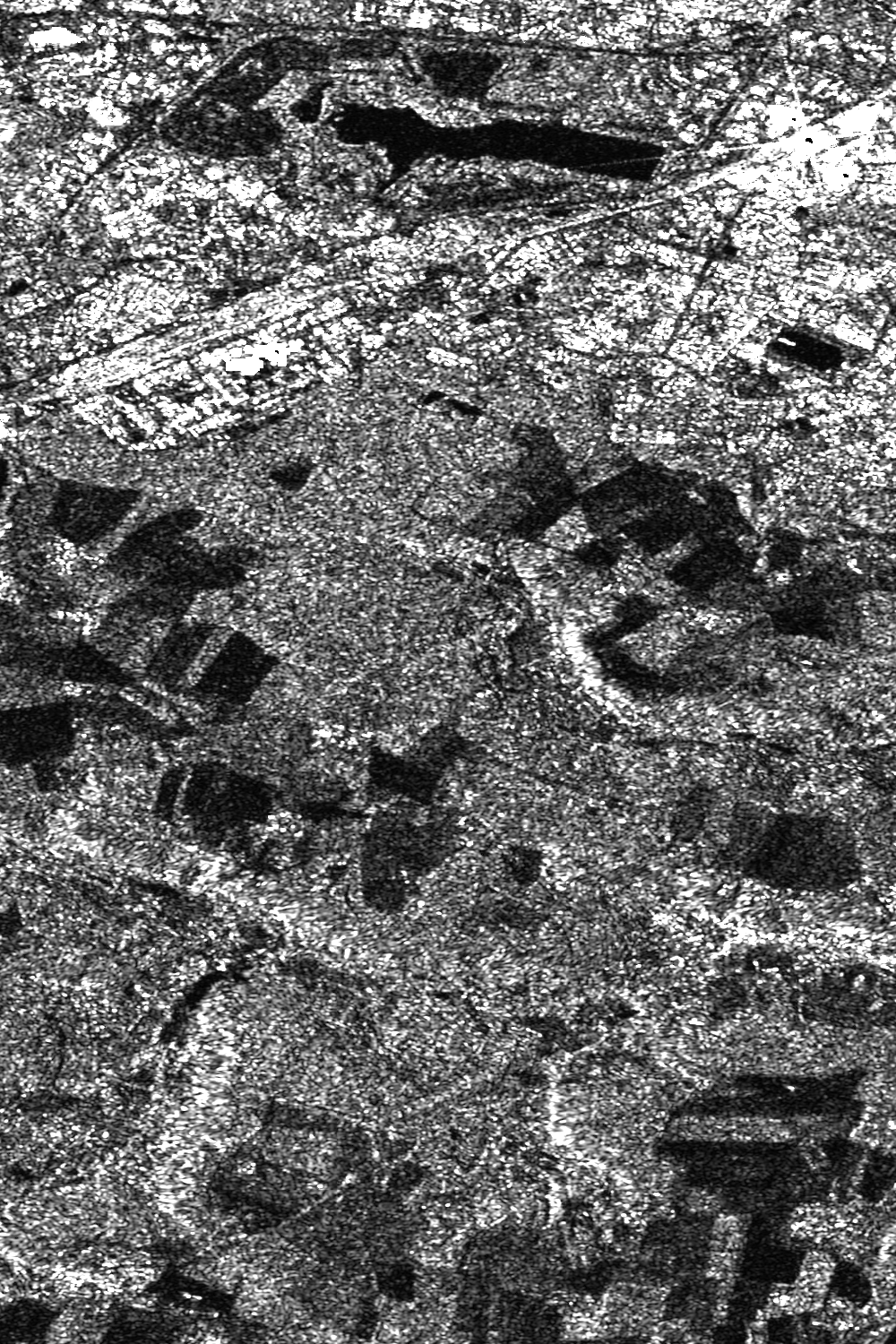} \\
    \textbf{(a)} & \textbf{(b)} & \textbf{(c)}\\
    \end{tabular}}
    \caption{Qualitative results on a real GRD image: \textbf{(a)} noisy input, \textbf{(b)} proposed model output and \textbf{(c)} SAR-BM3D output.}
    \label{fig:comparisions}
\end{figure*}

Also in this case our model outperforms all the others, as shown in Table \ref{tab:results2}, even if  with respect to the results presented in Table \ref{tab:enelquaresults}, run on our dataset, now the difference between the proposed model and the others is smaller. This outcome was expected as by definition the model must perform better on data similar to the ones used for training rather than on other data. Qualitative results are shown in Fig. \ref{fig:comparisions}, where the input noisy image, the proposed model prediction and the output of the  SAR-BM3D filter, are reported.  

\begin{table}[!ht]
    \centering
    \begin{tabular}{lc@{\hskip 1cm}lc}
    \toprule
    \textbf{Model} & \textbf{ENL} & \textbf{Model} & \textbf{ENL}\\
    \midrule
    \textbf{Proposed} & \textbf{69.72} & 
    \textbf{Mean}& 31.24\\
    
    \textbf{Lee} & 32.19 &
    \textbf{Median} & 31.12\\
    
    \textbf{Lee Enhanced} & 34.02 & 
    \textbf{Fastnl} & 25.31\\
     
    \textbf{Kuan}& 32.45 &
    \textbf{Bilateral} & 32.48\\
    
    \textbf{Frost} & 33.80 &
    \textbf{SAR-BM3D} & 45.39\\
    \bottomrule

    \end{tabular}
    \caption{Models' ENL computed on the real GRD image. Input with speckle,  Proposed Model Prediction,  Lee,  Lee Enhanced,  Kuan, Frost, Mean,  Median,  Fastnl, Bilateral, SAR-BM3D}
    \label{tab:results2}
\end{table}

To conclude this experimental section, Table \ref{tab:processing_time} presents the processing time of two filters applied to the real GRD image, the proposed one and the SAR-BM3D which in the last experiments produces closer results. Both the time needed to process a patch of $96\times 96$ pixels and the full size image of $10.000\times 10.000$ pixels are shown. The machine used to run these tests is a MacBook Pro 13 with an Intel(R) Core(TM) i5-8279U CPU @2.40 GHz and 7.85 GB of available RAM. From the Table \ref{tab:processing_time} it is evident that one additional advantage of the proposed model is the processing time reduction. 

\begin{table}[!ht]
    \centering
    \begin{tabular}{lccc}
        \toprule
         Filter & Time per patch & Total time  & Software\\
         \midrule
         SAR-BM3D      & $\sim$ 2 s   & $\sim$ 6 h & MatLab\\
         \textbf{Proposed} & $\mathbf{\sim}$ \textbf{0.08 s} & $\mathbf{\sim}$ \textbf{13 m} & \textbf{Python}\\
         \bottomrule
    \end{tabular}
    \caption{Processing time for the Proposed model and SAR-BM3D.}
    \label{tab:processing_time}
\end{table}

\section{Further analysis}
In this section further experiments and analysis are presented to give the reader some insights about edges preservation and statistical characteristics preservation for the proposed method.

\subsection{Edges Preservation}
To evaluate the edge preservation of the proposed method, we applied the Sobel operator. This operator, commonly used in image processing and computer vision, is able to create images with edge emphasized. It  uses two kernels, one for the horizontal and one for the vertical axis, which are convolved with the image to calculate approximately its gradient \cite{kanopoulos1988design}.

Let $Y$ be a SAR image filtered by the proposed method, then the horizontal changes $H_x$ are calculated by:

\begin{equation}
    \centering
    H_x = \begin{bmatrix}
    -1 & 0 & +1\\
    -2 & 0 & +2\\
    -1 & 0 & +1
   \end{bmatrix}*Y
\end{equation}

while the vertical changes $H_y$ are calculated by:

\begin{equation}
    \centering
    H_x = \begin{bmatrix}
    -1 & -2 & -1\\
    0 & 0 & 0\\
    +1 & +2 & +1
   \end{bmatrix}*Y
\end{equation}

The image gradient is then obtained by $H=\sqrt{H^2_x+H^2_y}$.

\begin{figure}[!ht]
    \centering
    \includegraphics[width=1.0\columnwidth]{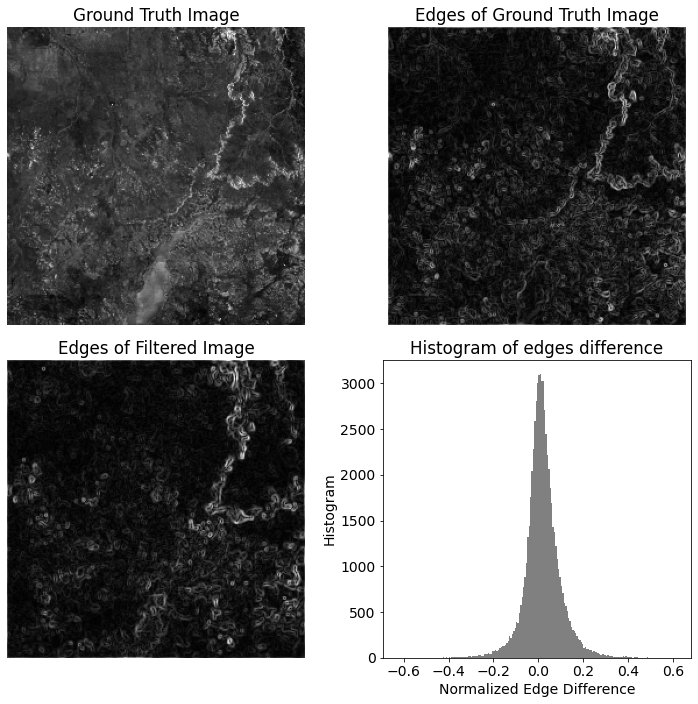}
    \caption{Top row:  Input Image without speckle, edges derived through the Sobel operator. Bottom row: edges of model prediction made on the speckled version of the input image, and the histogram of the differences between edges in the two cases.}
    \label{fig:edge1}
\end{figure}

\begin{figure}[!ht]
    \centering
    \includegraphics[width=1.0\columnwidth]{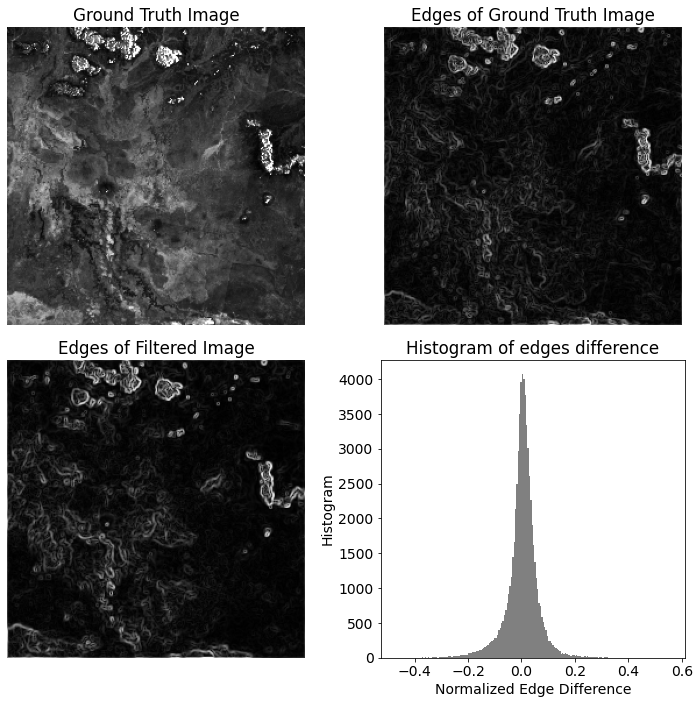}
    \caption{Edge preservation evaluation as Figure 8 on a different image.}
    \label{fig:edge2}
\end{figure}

Figure \ref{fig:edge1} and \ref{fig:edge2} report two examples of edge preservation evaluation for our method. These figures are structured in two rows: the first row shows the speckle free image and an image with the edges obtained by applying the Sobel operator; the second row shows the edges obtained by applying the Sobel operator to the image filtered by using our method and the histogram of the difference between the two results. 
It is possible to note that the proposed method is able to preserve most of the edges, with a little loss. Indeed, as shown in the histogram, the distribution of the difference between the input image edges' values and the predicted image edges' values is centered in zero with a maximum of 0.4 for a limited number of pixels.
It is important to highlight that an important role in the image edges preservation is played by the designed loss function that is based on the SSIM. This element is in fact responsible of forcing the model, during the training phase, in producing a filtered image with a high structural similarity with respect to the ground truth. This characteristic has an impact not only on  the edge preservation but also on the spatial information preservation, that is also a crucial aspect.

\subsection{Preservation of the statistical characteristics}
In order to measure if the input distribution is preserved, we compared the distribution of a ground truth image with respect to the filtered one. Starting from the ground truth image as obtained through the temporal average of a long times series of many images, we calculated the histogram and the estimated probability distribution function (pdf).
\begin{figure}[!ht]
    \centering
    \includegraphics[width=0.9\columnwidth]{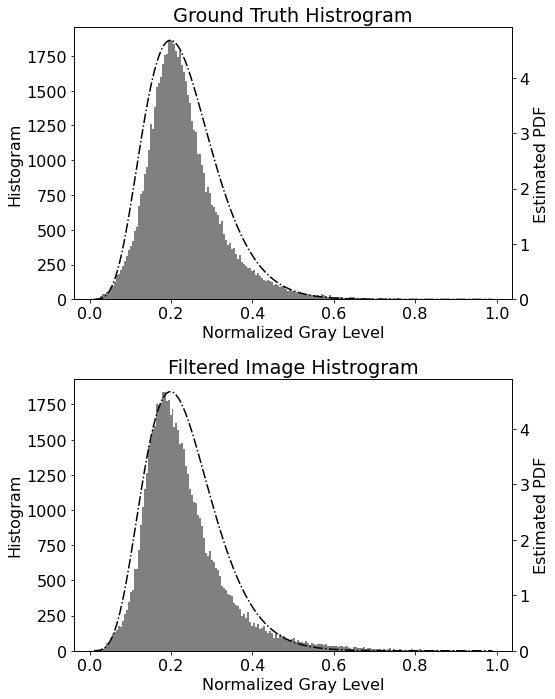}
    \caption{Example of statistical characteristics preservation}
    \label{fig:stats3}
\end{figure}
Then, by simulating the speckle as specified in Section \ref{sec:dataset}, we obtained the input image for our model, and by filtering it, the output image. The histogram of this latter image has been also calculated.  Figure \ref{fig:stats3} shows the comparison between the histogram of the filtered image against the histogram of the ground truth.  The same pdf is reported for both, and it has been calculated by fitting the ground truth data with a Gamma distribution as specified in \cite{moser2006sar}. 
We left the pdf of the ground truth image also for the output image to show that although there are variations (for example the mean) in the filtered image, this latter falls in the same pdf of the input image, thus demonstrating that the statistical characteristics are preserved by the proposed filter.
It is important to underline, that our method does not force the preservation of the mean and the variance, but forces the predicted distribution to be as much as possible equal to the distribution of the ground truth. Thus, there could be cases where the mean is not perfectly preserved, but since the distribution is forced to be the same, the statistical properties of the filtered images are saved to some extent. In our opinion this is an important result, since  regarding the mean preservation we can always think about methodological approaches to reach this goal. 

\section{Conclusions}
In this paper, a method for SAR speckle filtering, based on residual CNNs, has been presented and discussed. 

The main idea is to use the result of a very long time-series with many images, a time-averaged cumulative smoothing of the sample area images in the GEE as a multiplicative noise-free image. Another important novelty refers to the use of a subtractive layer in a residual CNN with not log-scaled GEE GRD High Resolution SAR intensities data.  

Several examples of applications based on the employment of GRD SAR data have been given, by underlining how  their use has greatly increased over the years due to lower computational load with respect to other SAR products, and due to their availability on cloud platforms like GEE.

Both the dataset and the model have been developed from scratch and made available (with the code open-source) in the two versions, time-averaged cumulative smoothed intensities and time-averaged cumulative smoothed amplitudes, on Git-Hub for further analysis and investigation to allow interested researchers to use what best fits the specific case study and application. 

The authors have made their best to realize a code easy to read, and they believe in the principles of open access policy and sharing research outputs content in accordance to the inspiring principles of Professor Landgrebe, whose studies have been  instrumental in the inception and growth of spectral RS image analysis.

Experiments have been first run on the COCO dataset to prove the generalization of the proposed model, that can be also used for noise filtering of common images, as proven by the results. Then, other experiments have been run on the generated dataset and on a real GRD image, showing  that the proposed method works successfully in filtering the speckle from the SAR images. The approach as described leads  to better results by a significant margin with respect to all the other methods found in the literature, both on the proposed dataset and on the real GRD image.

The performance can still be improved, for instance through a further tuning of the network hyperparameters, an optimal selection of the number of layers and loss function weights, that might help in reducing both the amount of noise and the image blurring.

Among the future developments, there might be the integration of a much more complex speckle model, adaptable to the typology of the terrain and land cover. Furthermore, the model might be trained on a dataset containing speckled images  with different SNR values, in order to make it more robust \cite{he2019parametric}. 

Another important aspect that might be better investigated is the eventual impact of phenomena such as the humidity of the ground surface and human moving targets on the generation of the dataset. 

\appendices
\section{Speckle Model}\label{app:speckle}
As highlighted in the manuscript, the proposed dataset has been made available online  open access \cite{dataset}. It has been released in two versions, time-averaged cumulative smoothed intensities and time-averaged cumulative smoothed amplitudes. In the two cases, we report in this appendix the statistical models used to generate the images with the speckle.

The most common and widely used statistical model used to represent SAR speckle for distributed scatters is the multiplicative noise model, as proposed by Goodman  \cite{goodman1976some}:

\begin{equation}
    \centering
    Y = X\cdot S
\end{equation}

where $Y \in \mathbb{R}^{W\times H}$ is the observed intensity acquisition, $X \in \mathbb{R}^{W\times H}$ is the noise-free version of the acquisition and $S \in \mathbb{R}^{W\times H}$ is the speckle noise.  By considering a multi-looking (L looks) SAR acquisition, $S$ follows the Gamma distribution with unitary mean and variance $1/L$ \cite{article_book_goodman}. The corresponding probability density function is defined by:

\begin{equation}
    \centering
    f(S) = \frac{L^L S^{L-1} e^{-LS}}{\Gamma(L)},\ \ S\geq 0, L\geq 1
\end{equation}

where $\Gamma$ represents the Gamma function. \\The same model can be applied to the amplitude image, square root of the intensity:

\begin{equation}
    \centering
    y = x \cdot s
\end{equation}

where $y \in \mathbb{R}^{W\times H}$ is the observed amplitude acquisition, $x \in \mathbb{R}^{W\times H}$ is the speckle-free amplitude image and $s \in \mathbb{R}^{W\times H}$ is the speckle noise.

In this case $s$ follows the Nakagami distribution, with a probability density function defined by the following equation:

\begin{equation}
    \centering
    f(s) = \frac{2L^L s^{2L-1} e^{-Ls^2}}{\Gamma(L)},\ \ S\geq 0, L\geq 1
\end{equation}

that corresponds to the Rayleigh distribution for  $L=1$ (single-look) \cite{lattari2019deep}, \cite{article_book_goodman}.

\bibliographystyle{IEEEtran}
\bibliography{main}

\begin{IEEEbiography}[{\includegraphics[width=1in,height=1.25in,clip,keepaspectratio]{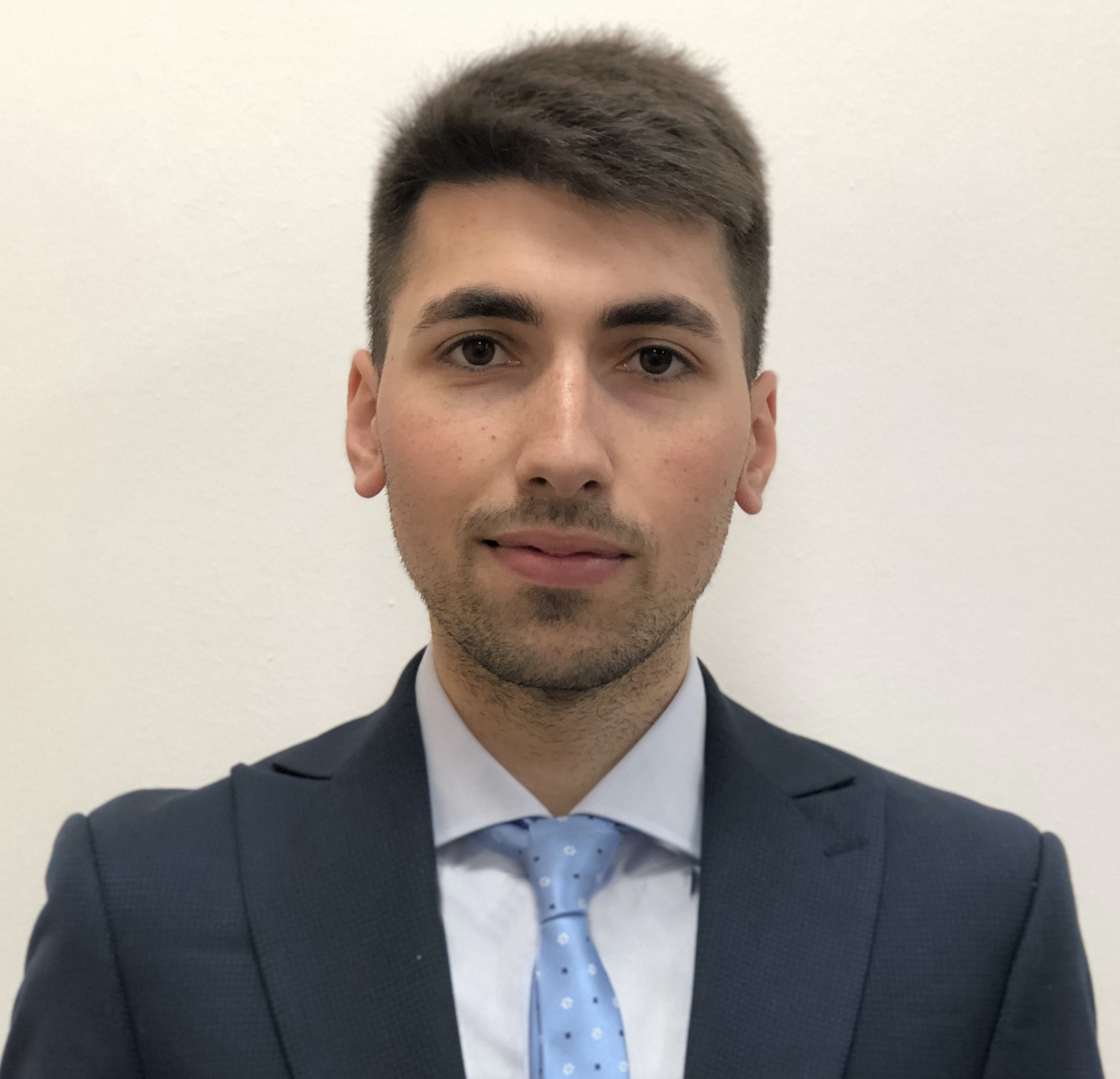}}]{Alessandro Sebastianelli} graduated  with laude in Electronic Engineering for Automation and Telecommunications at the University  of Sannio in 2019. He is enrolled in the Ph.D. program with University of Sannio, and his research topics mainly focus on Remote Sensing and Satellite data analysis, Artificial  Intelligence  techniques for Earth Observation, and data fusion. He has co-authored several papers in reputed journals and conferences for  the  sector  of  Remote Sensing. He has been a visited researcher at Phi-lab in European Space  Research  Institute  (ESRIN)  of  the  European  Space Agency (ESA), in Frascati, and still collaborates with the Phi-lab  on  topics  related  to  deep  learning  applied  to  Earth Observation. He has won an ESA OSIP proposal in August 2020 presented with his Ph.D. Supervisor, Prof. Silvia L. Ullo.
\vspace{-1.3cm}

\end{IEEEbiography}
\begin{IEEEbiography}[{\includegraphics[width=1in,height=1.25in,clip,keepaspectratio]{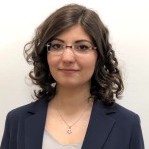}}]{Maria Pia Del Rosso} graduated  with  laude  in  Electronics Engineering for Automation and Telecommunications  at  the  University  of Sannio  in  October 2019 and she is currently a PhD candidate.  As a master student, she has been a visiting researcher at the Phi-lab in the European Space  Research  Institute  (ESRIN)  of  the  European  Space Agency (ESA), in Frascati. She worked on applying Deep Learning techniques to Remote Sensing Earth Observation data for monitoring geohazard phenomena, such as earthquakes, landslides and volcanic eruptions. She is currently working on the matching between multispectral and radar data applying AI techniques.
\end{IEEEbiography}

\vspace{-0.3cm}

\begin{IEEEbiography}[{\includegraphics[width=1in,height=1.15in,clip,keepaspectratio]{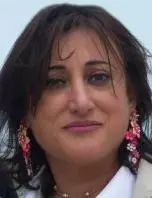}}]{Silvia Liberata Ullo} IEEE Senior Member, Industry Liaison for IEEE Joint ComSoc/VTS Italy Chapter. National Referent for FIDAPA BPW Italy Science and Technology Task Force. Researcher since 2004 in the Engineering Department of the University of Sannio, Benevento (Italy). Member of the Academic Senate and the PhD Professors’ Board. She is teaching: Signal theory and elaboration, and Telecommunication networks for Electronic Engineering, and Optical and radar remote sensing for the Ph.D. course. Authored 80+ research papers, co-authored many book chapters and served as editor of two books, and many special issues in reputed journals of her research sectors. Main interests: signal processing, remote sensing, satellite data analysis, machine learning and quantum ML, radar systems, sensor networks, and smart grids. Graduated with Laude in 1989 in Electronic Engineering,  at the Faculty of Engineering at the Federico  II University, in  Naples, she pursued the M.Sc. degree from the Massachusetts Institute  of Technology (MIT) Sloan  Business  School  of  Boston,  USA,  in June 1992.  She has worked in the private and public sector from 1992 to 2004, before joining the University of Sannio.
\end{IEEEbiography}

\begin{IEEEbiography}[{\includegraphics[width=1in,height=1.15in,clip,keepaspectratio]{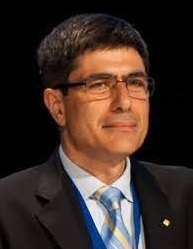}}]{Paolo Gamba}
IEEE Fellow, received the Laurea (cum laude) and Ph.D. degrees in electronic engineering from the University of Pavia, Pavia, Italy, in 1989 and 1993, respectively.
He is a Professor of telecommunications with the University of Pavia, where he leads the Telecommunications and Remote Sensing Laboratory and serves as a Deputy Coordinator of the Ph.D. School in Electronics and Computer Science. He has been invited to give keynote lectures and tutorials in several occasions about urban remote sensing, data fusion, EO data, and risk management.
Dr. Gamba has served as the Chair for the Data Fusion Committee of the IEEE Geoscience and Remote Sensing Society from 2005 to 2009. He has been elected in the GRSS AdCom since 2014. He is also the GRSS President. He had been the Organizer and Technical Chair of the biennial GRSS/ISPRS Joint Workshops on Remote Sensing and Data Fusion over Urban Areas from 2001 to 2015. He has also served as the Technical Co-Chair of the 2010, 2015, and 2020 IGARSS Conferences, Honolulu, HI, USA, and Milan, Italy, respectively. He was the Editor-in-Chief of the IEEE Geoscience and Remote Sensing Letters  from 2009 to 2013.
\end{IEEEbiography}
\end{document}